\documentclass[10pt,twocolumn,letterpaper]{article}

\PassOptionsToPackage{table}{xcolor}

\usepackage[pagenumbers]{cvpr}

\usepackage{microtype}

\usepackage{graphicx}
\usepackage{booktabs}
\usepackage{colortbl}
\usepackage{multirow}
\usepackage{enumitem}
\usepackage{algorithm}
\usepackage{algpseudocode}
\usepackage[T1]{fontenc}

\definecolor{tabfirstcolor}{RGB}{255,179,179}
\definecolor{tabsecondcolor}{RGB}{255,217,179}
\newcommand{\tabfirst}[1]{\cellcolor{tabfirstcolor}#1}
\newcommand{\tabsecond}[1]{\cellcolor{tabsecondcolor}#1}

\definecolor{cvprblue}{rgb}{0.21,0.49,0.74}
\usepackage[pagebackref,breaklinks,colorlinks,allcolors=cvprblue]{hyperref}

\def\confName{CVPR}
\def\confYear{2027}

\title{SprintGS: Eliminating Redundancy for Fast 3D Gaussian Splatting Training}

\author{Ryosuke Watanabe \hspace{1cm} Tomoaki Konno \\
KDDI Research, Inc.
\\
\texttt{\{ru-watanabe,tm-konno\}@kddi.com}}

\begin{document}
\maketitle

\begin{abstract}
3D Gaussian Splatting (3DGS) renders photorealistic novel views in real time, but its per-scene training (optimization) remains slow. Recent fast training methods such as FastGS shorten training to a few minutes by reducing the number of Gaussians.
Profiling FastGS, we find two costs that remain even in its compact model. First, rasterization is still the largest cost, and redundant work remains in it. Second, the loss computation does not shrink with the number of Gaussians, so its share of the training time grows as the model becomes compact. We present SprintGS, a training framework built on FastGS that keeps the model compact and reduces both costs. For rasterization, \textbf{scheduled tile culling} limits each Gaussian to the tiles where it matters, using a coarse footprint that is annealed to the exact one before training ends; \textbf{convergence-aware pruning} removes converged, nearly transparent Gaussians; and a \textbf{shuffle-efficient backward pass} reduces the data exchanged between GPU threads. For the loss, a \textbf{fused loss} computes the loss and its image gradient in the forward pass without a loss map, removing most of its cost. SprintGS trains $2.15\times$ faster than FastGS at comparable quality, and $1.81\times$ faster in its high-quality setting. It also accelerates other fast training methods and dynamic, sparse-view, and surface reconstruction.
\end{abstract}

\section{Introduction}
\label{sec:intro}

Novel view synthesis (NVS), which reconstructs a scene from a set of photographs and renders it from unseen viewpoints, is a long-standing problem in computer vision and graphics.
3D Gaussian Splatting (3DGS)~\cite{Vanilla3DGS} has become a popular approach to this task: it represents a scene by anisotropic 3D Gaussians and rasterizes them into photorealistic images at interactive frame rates.
However, 3DGS requires a per-scene optimization before any image can be rendered.
Although recent methods have substantially shortened this optimization, it is still far from instantaneous.

Prior work speeds up this optimization in two ways.
The first optimizes the implementation of training, from GPU kernels to framework overhead, and trains a model with as many Gaussians as 3DGS in a short time~\cite{FasterGS,gsplat,ImprovedGS+,DISTWAR}.
The second reduces the number of Gaussians by revising how they are added (densification) and removed (pruning)~\cite{Taming3DGS,SpeedySplat,DashGaussian,FastGS}.
FastGS~\cite{FastGS}, for example, densifies and prunes Gaussians according to their contribution
to multi-view reconstruction quality, and trains a scene with far fewer Gaussians than 3DGS.
Such a compact model also reduces storage, GPU memory, and rendering time,
which matters on limited hardware and for transmitting scenes.
Even so, a user who wants to check a captured scene on the spot must still wait minutes for the trained model, although rendering runs in real time.
This paper aims to further shorten this wait for compact models while keeping their quality.

\begin{figure}[t]
  \centering
  \includegraphics[width=\linewidth]{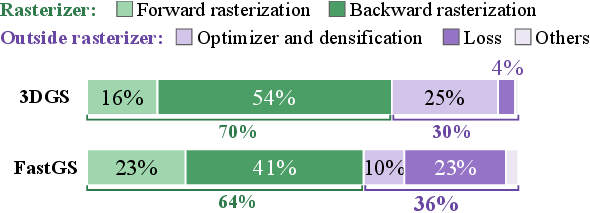}
  \caption{Training-time breakdown of 3DGS~\cite{Vanilla3DGS} and FastGS~\cite{FastGS} over the 13 scenes of Mip-NeRF 360~\cite{MipNeRF360}, Deep Blending~\cite{DeepBlending}, and Tanks \& Temples~\cite{TT}.}
  \label{fig:breakdown}
\end{figure}

\begin{figure}[t]
  \centering
  \includegraphics[width=\linewidth]{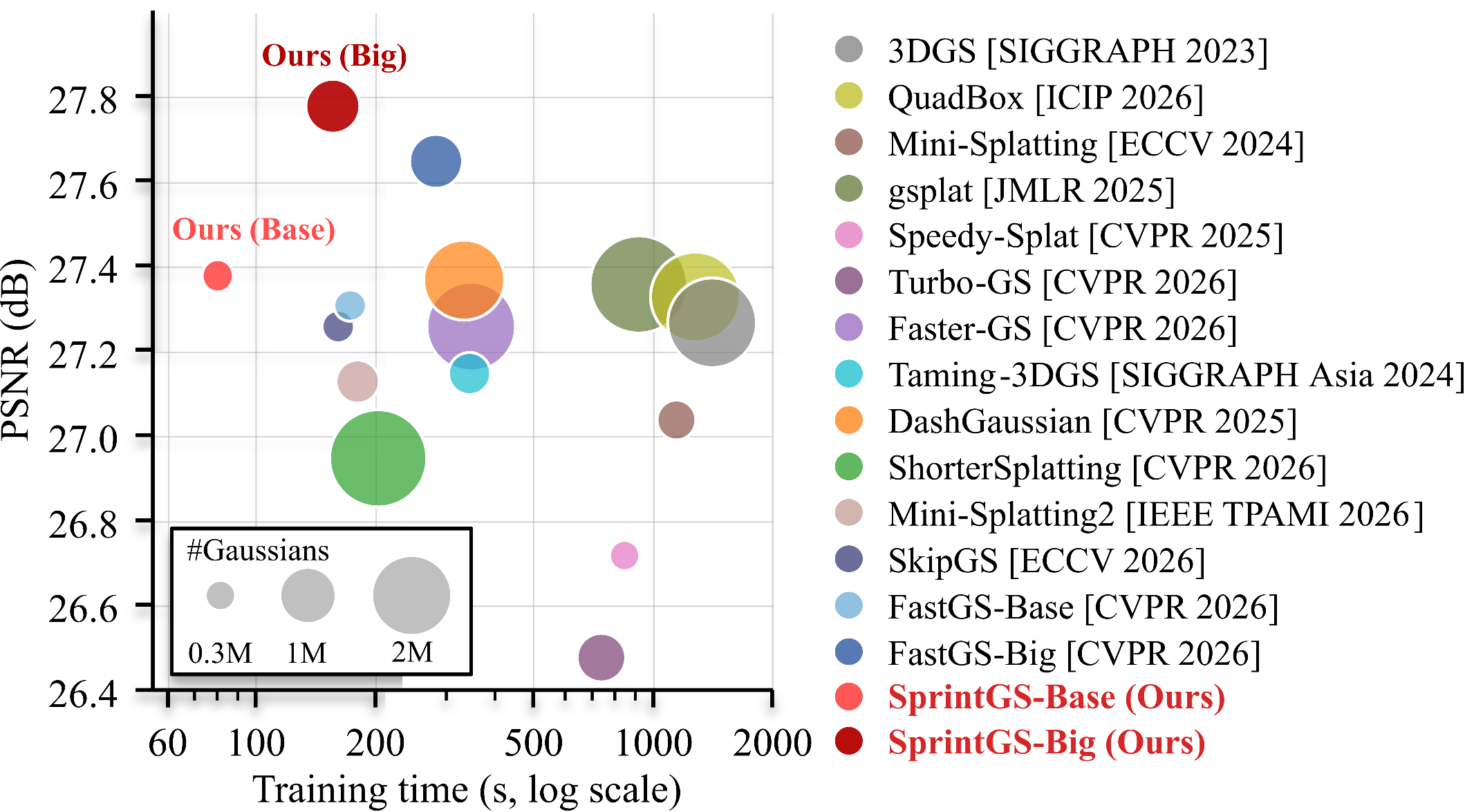}
  \caption{Training time and PSNR of fast 3DGS training methods, averaged over the 13 scenes of Mip-NeRF 360~\cite{MipNeRF360}, Deep Blending~\cite{DeepBlending}, and Tanks \& Temples~\cite{TT}.
  The marker area is proportional to the number of Gaussians.}
  \label{fig:teaser}
\end{figure}

To find where the training time of a compact model goes, we profile the training of 3DGS~\cite{Vanilla3DGS} and FastGS~\cite{FastGS}~(\cref{fig:breakdown}).
We find that rasterization is still the largest cost of a compact model, and that the share of the loss computation grows as the model becomes compact.
Our strategy is therefore to keep the model compact and reduce the costs of both rasterization and the loss.

Rasterization takes 64\% of the training time of FastGS.
Although FastGS reduces the number of Gaussians, redundant work remains in three places:
each Gaussian is rasterized over many tiles where it contributes almost nothing to the image; some Gaussians have converged and become nearly transparent but stay in the model;
and the backward pass moves more per-pixel data between GPU threads than the gradients require.
In contrast, the loss is computed per pixel, so its cost does not shrink with the number of Gaussians.
While the share of the optimizer and densification falls from 25\% to 10\%,
the loss takes a similar time in both methods, and its share grows from 4\% to 23\%.

We thus present SprintGS, a training framework built on FastGS.
Our contributions are as follows:
\begin{itemize}
  \item \textbf{Removing redundancy in rasterization.}
  We propose three components, one for each of the three places above.
  \textit{Scheduled tile culling} limits each Gaussian to the tiles where it matters, using a coarse footprint that is annealed to the exact one before training ends. \textit{Convergence-aware pruning} removes converged, nearly transparent Gaussians early and reallocates part of the freed budget to poorly reconstructed regions. A \textit{shuffle-efficient backward pass} reduces the data exchanged between GPU threads.
  \item \textbf{Reducing the loss cost.}
  We identify the loss computation as a major cost of fast training and remove most of it with a \textit{fused loss}, which computes the loss and its image gradient in the forward pass without a loss map.
  \item \textbf{Speed-up at comparable quality.}
    SprintGS-Base trains $2.15\times$ faster than FastGS-Base~\cite{FastGS} at comparable quality (\cref{fig:teaser}; with optimizations from Faster-GS~\cite{FasterGS}) and also accelerates other fast training methods and tasks.
\end{itemize}

\section{Related Work}
\label{sec:related-work}
Since 3DGS~\cite{Vanilla3DGS} takes tens of minutes to train a scene, many methods have been proposed to shorten its training. We review them below, grouped by the source of the speed-up.

\subsection{Reducing the Number of Gaussians}
A key bottleneck of 3DGS training is the excessive number of Gaussians.
Many methods~\cite{ImprovedGS,3DGS-MCMC,SteepGS,mini-splating,mini-splating2,SpeedySplat,SAD-GS,Taming3DGS,Color-cued-densification,FastGS,TurboGS_CVPR2026,GroupTraining_ICCV2025,POP_ECCV2026}
refine densification and pruning to limit the number of Gaussians.
Among them, FastGS~\cite{FastGS} is most related to ours: it densifies Gaussians that consistently cover high-error regions across views
and prunes those with little multi-view contribution.

\subsection{Reducing the Work in Rasterization}

\noindent\textbf{Gaussian--tile pairs.}
Tile-based rasterization assigns each Gaussian to every tile its 2D footprint may overlap,
and both the forward and backward passes process the resulting \emph{Gaussian--tile pairs}.
3DGS~\cite{Vanilla3DGS} bounds each footprint by a square that overestimates it, producing many redundant pairs.
To accelerate rendering, several renderers remove such pairs with tighter Gaussian--tile tests~\cite{StopThePop,FlashGS}.
Others cull Gaussians hierarchically~\cite{HiGS} or per warp~\cite{LocalGS}.
Since every training iteration runs the rasterizer in both passes, fewer pairs also shorten training.
Speedy-Splat~\cite{SpeedySplat} assigns each Gaussian exactly to the tiles its footprint ellipse intersects,
and other methods use tighter footprints during optimization~\cite{ADRGaussian,QuadBox,FastGS,gsplat}
or shrink the Gaussians so that fewer of them overlap each pixel~\cite{ShorterSplatting}.

\noindent\textbf{Pixels.}
EAGLES~\cite{EAGLES} and DashGaussian~\cite{DashGaussian} render at reduced resolution early in training, and TurboGS~\cite{TurboGS_ICML2026} optimizes on sparse pixels sampled by reconstruction error.

\noindent\textbf{Converged parts.}
Much of the computation is spent on parts that have already converged. SkipGS~\cite{SkipGS} skips the backward pass of views whose loss has stabilized, and GSta~\cite{GSta} freezes Gaussians with small gradients.

\noindent\textbf{Backward pass.}
The original implementation~\cite{Vanilla3DGS} parallelizes the backward pass over pixels, so many threads add gradients to the same Gaussian with contended atomics, which DISTWAR~\cite{DISTWAR} mitigates with warp-level reduction; Balanced 3DGS~\cite{Balanced3DGS} balances the rasterization workload during training. Taming-3DGS~\cite{Taming3DGS} instead parallelizes over Gaussians: each thread of a warp handles one Gaussian, and per-pixel states pass from thread to thread in depth order.
This splat-parallel backward pass is the backbone of recent fast trainers~\cite{FastGS,DashGaussian,FasterGS}.

\subsection{Reducing the Cost outside the Rasterizer}
Each iteration also computes the loss, updates the parameters, and synchronizes the CPU with the GPU.
Several methods reduce these costs at the implementation level:
Taming-3DGS~\cite{Taming3DGS} computes SSIM with a fused kernel,
Faster-GS~\cite{FasterGS} fuses the activations and the optimizer steps among many other optimizations,
and ImprovedGS+~\cite{ImprovedGS+} reduces host--device synchronization.
Other work distributes training over multiple GPUs~\cite{Grendel_ICLR2025}.

\subsection{Accelerating Convergence and Beyond}
Dense initializations from foundation models or correspondences~\cite{EDGS,FFInit,QuickSplat_ICCV}
and second-order optimizers~\cite{3DGS-LM,3DGSSquared,3DGS2-TR} reduce the number of iterations, the latter at a higher cost per iteration. 
Feed-forward networks~\cite{pixelSplat,MVSplat_ECCV2024,Anysplat,ReSplat_ECCV2026} avoid per-scene optimization
through large-scale pretraining;
we instead accelerate per-scene optimization, which remains widely used for high-fidelity reconstruction.

\begin{figure*}[t]
  \centering
  \includegraphics[width=\textwidth]{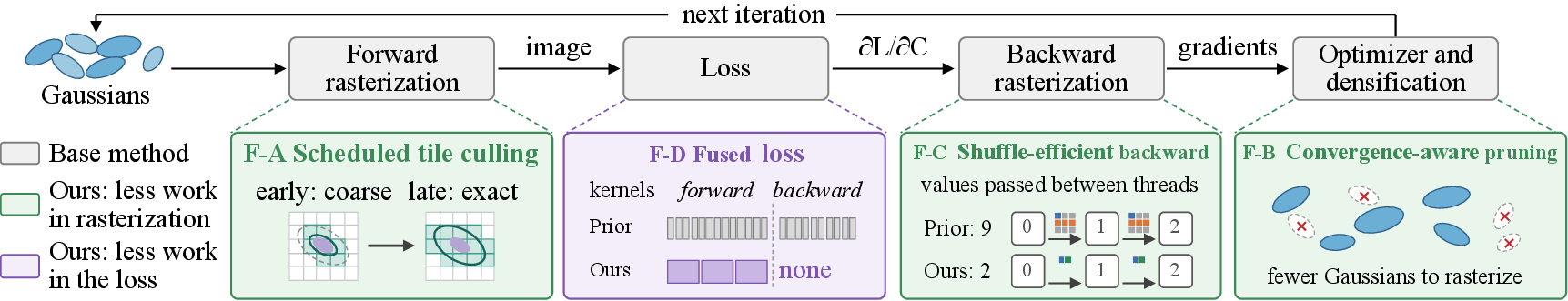}
  \caption{Overview of SprintGS.
  Gray boxes show one training iteration of the base method.
  Each colored box is a close-up of the step in which a component runs.
  F-A, F-B, and F-C (green) reduce redundant work in rasterization.
  F-D (purple) reduces the cost of the loss by computing the loss and its image gradient in the forward pass.
  Because these steps are shared by 3DGS-based methods,
  the components also apply to other fast training methods and other tasks
  (\cref{sec:other_baselines,sec:extension}).}
  \label{fig:overview}
\end{figure*}

\section{Preliminaries}
\label{sec:prelim}

\noindent \textbf{3D Gaussian Splatting.}
3DGS~\cite{Vanilla3DGS} represents a scene by 3D Gaussians.
The $i$-th Gaussian has a position (mean) $\boldsymbol{\mu}_i\in\mathbb{R}^3$, a covariance $\boldsymbol{\Sigma}_i=\mathbf{R}_i\mathbf{S}_i\mathbf{S}_i^\top\mathbf{R}_i^\top$ given by a rotation $\mathbf{R}_i$ and a scale $\mathbf{S}_i$, an opacity $o_i\in(0,1)$, and spherical harmonics coefficients of a view-dependent color $\mathbf{c}_i$.
For rendering, each Gaussian is projected to a 2D Gaussian with mean $\boldsymbol{\mu}'_i$ and covariance $\boldsymbol{\Sigma}'_i=\mathbf{J}\mathbf{W}\boldsymbol{\Sigma}_i\mathbf{W}^\top\mathbf{J}^\top$, where $\mathbf{W}$ and $\mathbf{J}$ are the viewing transform and the Jacobian of the local affine approximation of the projection, respectively.
Its alpha value at a pixel $\mathbf{p}$ is  % [CHANGED] contribution -> alpha value
\begin{equation}
  \alpha_i=o_i\exp\bigl(-d_i^2/2\bigr),
  \label{eq:alpha}
\end{equation}
where $d_i^2=(\mathbf{p}-\boldsymbol{\mu}'_i)^\top\boldsymbol{\Sigma}'^{-1}_i(\mathbf{p}-\boldsymbol{\mu}'_i)$
is the squared Mahalanobis distance.
Let the $N$ Gaussians that overlap the pixel be indexed by $i=1,\dots,N$ from front to back.
The pixel color is
\begin{equation}
  \mathbf{C}=\sum_{i=1}^{N} \mathbf{c}_i\,\alpha_i\,T_i + T_{\mathrm{fin}}\,\mathbf{b},
  \qquad T_i=\prod_{j=1}^{i-1}(1-\alpha_j),
  \label{eq:render}
\end{equation}
where $T_i$ is the transmittance (the fraction of light reaching the $i$-th Gaussian) and $T_{\mathrm{fin}}=\prod_{j=1}^{N}(1-\alpha_j)$ is the final transmittance after all $N$ Gaussians,
and $\mathbf{b}$ is the background color.
Starting from an initial point cloud, typically obtained by structure-from-motion~\cite{COLMAP}, the Gaussians are optimized with Adam~\cite{Adam} for $T=30{,}000$ iterations to minimize the mean $L=\sum_p \ell_p / M$ of
\begin{equation}
  \ell_p=(1-\lambda)\lvert C_p-C^{\mathrm{gt}}_p\rvert+\lambda(1-\mathrm{SSIM}_p),
  \label{eq:loss}
\end{equation}
where $C_p$, $C^{\mathrm{gt}}_p$, and $\mathrm{SSIM}_p$ are the rendered value, the ground-truth value, and the SSIM~\cite{SSIM} of the $p$-th color value of a training image, which has $M$ color values (three per pixel), and $\lambda=0.2$~\cite{Vanilla3DGS,FastGS}.
Meanwhile, adaptive density control~\cite{Vanilla3DGS} densifies and prunes the Gaussians.

\noindent \textbf{Tile-based rasterization.}
The rasterizer divides the image into tiles of $16\times16$ pixels.
In the forward pass, the pixels of each tile traverse its Gaussians front to back and skip
those with $\alpha_i<\epsilon=1/255$.
By \cref{eq:alpha}, a Gaussian therefore affects only the pixels inside its exact footprint, the ellipse whose squared Mahalanobis radius is $\delta_i^{\mathrm{full}}$:
\begin{equation}
  d_i^2\le\delta_i^{\mathrm{full}}=2\ln(o_i/\epsilon).
  \label{eq:full_footprint}
\end{equation}
3DGS assigns each Gaussian to every tile that a bounding square of this footprint overlaps, and Speedy-Splat~\cite{SpeedySplat} only to the tiles intersecting the ellipse; both passes process all resulting Gaussian--tile pairs, so the cost grows with their number.

\noindent \textbf{FastGS.}
FastGS~\cite{FastGS} keeps the number of Gaussians small by multi-view consistent densification and pruning (\cref{sec:related-work}).
Its rasterizer uses the tile assignment of Speedy-Splat~\cite{SpeedySplat} on a footprint uniformly shrunk to $d_i^2\le\beta\,\delta_i^{\mathrm{full}}$ ($\beta=0.5$ or $0.7$, per dataset) and the splat-parallel backward pass of Taming-3DGS~\cite{Taming3DGS}, which processes the Gaussians of a tile front to back in buckets of 32 (\cref{sec:method_backward}).
Unless stated otherwise, SprintGS keeps all settings of FastGS.

\section{Proposed Method (SprintGS)}
\label{sec:method}

SprintGS builds on FastGS with four components (\cref{fig:overview}) and implementation optimizations adapted from prior work.
F-A to F-C remove redundant rasterization work: scheduled tile culling (\textbf{F-A}, \cref{sec:method_scheduled-tile-culling}), convergence-aware pruning (\textbf{F-B}, \cref{sec:method_prune}), and the shuffle-efficient backward pass (\textbf{F-C}, \cref{sec:method_backward}).
F-D reduces the cost of the loss: the fused loss (\textbf{F-D}, \cref{sec:method_loss}).
Finally, we adapt implementation optimizations of Faster-GS~\cite{FasterGS} to the FastGS pipeline (\textbf{F-E}, \cref{sec:method_impl}).

%%% F-A
\subsection{Scheduled Tile Culling (F-A)}
\label{sec:method_scheduled-tile-culling}

\begin{figure}[t]
  \centering
  \includegraphics[width=\linewidth]{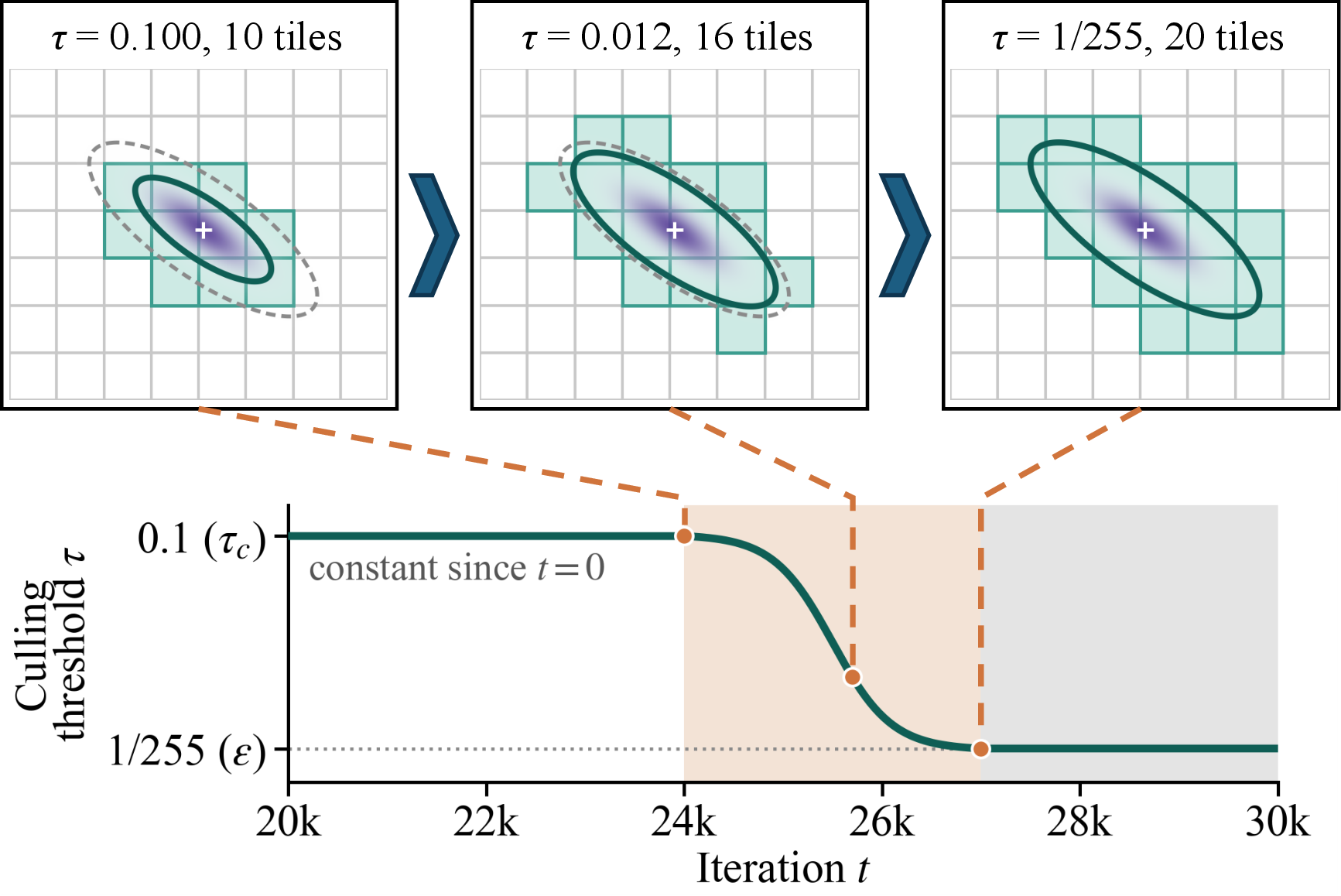}
  \caption{Scheduled tile culling.
  Top: a Gaussian projected onto the tile grid at three iterations. The solid ellipse is
  the footprint $d_i^2\le\delta_i(\tau)$ used for tile assignment, the dashed ellipse the exact footprint of Speedy-Splat~\cite{SpeedySplat} ($\tau=\epsilon$), and the shaded
  tiles the resulting Gaussian--tile pairs.
  Bottom: the culling threshold $\tau(t)$ of Eq.~\eqref{eq:tau_schedule}.
  $\tau$ is lowered from $\tau_c$ to $\epsilon$.}
  \label{fig:tile_culling}
\end{figure}

\noindent\textbf{Motivation.}
Existing tile assignments~\cite{SpeedySplat,FastGS} use the same footprint throughout training:
the exact one keeps all Gaussian--tile pairs,
whereas a uniformly shrunk one approximates the rendering even at test time.
We instead change the footprint \emph{over training}: a coarse footprint removes many pairs while the model is still far from converged,
and the exact one is restored before training ends.

\noindent\textbf{Tile culling.}
We replace the fixed threshold $\epsilon$ in \cref{eq:full_footprint} with a culling threshold $\tau(t)\ge\epsilon$ that \emph{varies over training},
and assign each Gaussian to the tiles intersecting the ellipse $d_i^2\le\delta_i(\tau)$, instead of the uniformly shrunk footprint of FastGS~\cite{FastGS}.
The size of this ellipse is that of the region $\alpha_i\ge\tau$, \ie, $\delta_i^{\mathrm{full}}$ with $\epsilon$ replaced by $\tau$, but not smaller than a lower bound:
\begin{equation}
  d_i^2\le\delta_i(\tau)=\max\bigl(2\ln(o_i/\tau),\ 0.15\,\delta_i^{\mathrm{full}}\bigr).
  \label{eq:alpha_contour}
\end{equation}
A larger $\tau$ drops more of the small alpha values below $\tau$ and assigns a Gaussian to fewer tiles
(\cref{fig:tile_culling}, top), while $\tau=\epsilon$ recovers the exact footprint of \cref{eq:full_footprint}.
The lower bound $0.15\,\delta_i^{\mathrm{full}}$ is needed because $o_i\le\tau$, \eg, after initialization or opacity reset, would otherwise give $\delta_i(\tau)\le0$, leaving the Gaussian on no tile and thus without gradients.

\noindent \textbf{Schedule.}
A large $\tau$ removes many pairs but biases the rendered image and its gradients,
whereas the exact footprint is unbiased but slow.
Since only the final model is evaluated, we use a coarse footprint for most of training
and the exact one at the end:
\begin{equation}
  \tau(t)=
  \begin{cases}
    \tau_c, & t<T-2\Delta,\\
    \tau_c\,(\epsilon/\tau_c)^{s(u)}, & T-2\Delta\le t<T-\Delta,\\
    \epsilon, & t\ge T-\Delta,
  \end{cases}
  \label{eq:tau_schedule}
\end{equation}
where $t$ is the iteration and $T$ is the number of iterations.
The transition lasts $\Delta=T/10$ iterations, $u=(t-T+2\Delta)/\Delta\in[0,1)$ is the progress within it,
and $s$ is a logistic function normalized to $s(0)=0$ and $s(1)=1$.
Thus $\tau$ decreases smoothly in log scale from $\tau_c$ to $\epsilon$
and stays at $\epsilon$ for the last $\Delta$ iterations (\cref{fig:tile_culling}, bottom).
The final model is also rendered with $\tau=\epsilon$ at test time, so the culling affects
only training.

%%% F-B
\subsection{Convergence-Aware Pruning (F-B)}
\label{sec:method_prune}

\noindent\textbf{Motivation.}
FastGS~\cite{FastGS} prunes Gaussians with little multi-view contribution only every 3,000
iterations after densification, so Gaussians that contribute little remain in the model
during densification and occupy part of the Gaussian budget.
F-B prunes them early and reallocates part of the freed budget to high-error regions, improving the quality--size trade-off.
Unlike pruning by scores computed over all training views after training~\cite{LightGaussian,PUP3DGS}
or freezing converged Gaussians~\cite{GSta},
F-B removes nearly transparent Gaussians during densification by reusing the optimizer state, without extra rendering.

\noindent\textbf{Convergence test.}
Let $v_i$ be the Adam second moment of the $i$-th Gaussian's position, summed over its three components,
\ie, an exponential moving average of squared positional gradients, used here as a convergence proxy.
During densification, we periodically prune the Gaussians
\begin{equation}
  \mathcal{P}=\bigl\{\,i \;\big|\; o_i<\theta_o\ \wedge\ v_i\le Q_q(v)\,\bigr\},
  \label{eq:fb_prune}
\end{equation}
where $\theta_o$ is an opacity threshold and $Q_q(v)$ is the $q$-quantile of $v$ over all Gaussians.
Neither condition suffices alone: a low opacity also marks Gaussians that may still become
opaque, and a small $v_i$ also marks opaque Gaussians that have converged to a useful state.
Their conjunction selects Gaussians that are nearly transparent and settled,
which are unlikely to contribute much to the final model.
We run the test every $K$ iterations from iteration 3,000 to the end of densification, skipping the $K$ iterations after each opacity reset.

\noindent\textbf{Budget re-allocation.}
The next time FastGS clones and splits Gaussians, we additionally clone or split the
$\lfloor\rho\,|\mathcal{P}|\rfloor$ Gaussians with the largest positional gradient among those
that FastGS marks as covering high-error regions but that fail its gradient threshold.
%Since $\rho<1$, the model remains smaller than without pruning in practice (\cref{tab:ablation}).

%%% F-C
\subsection{Shuffle-Efficient Backward Pass (F-C)}
\label{sec:method_backward}

\begin{figure}[t]
  \centering
  \includegraphics[width=\linewidth]{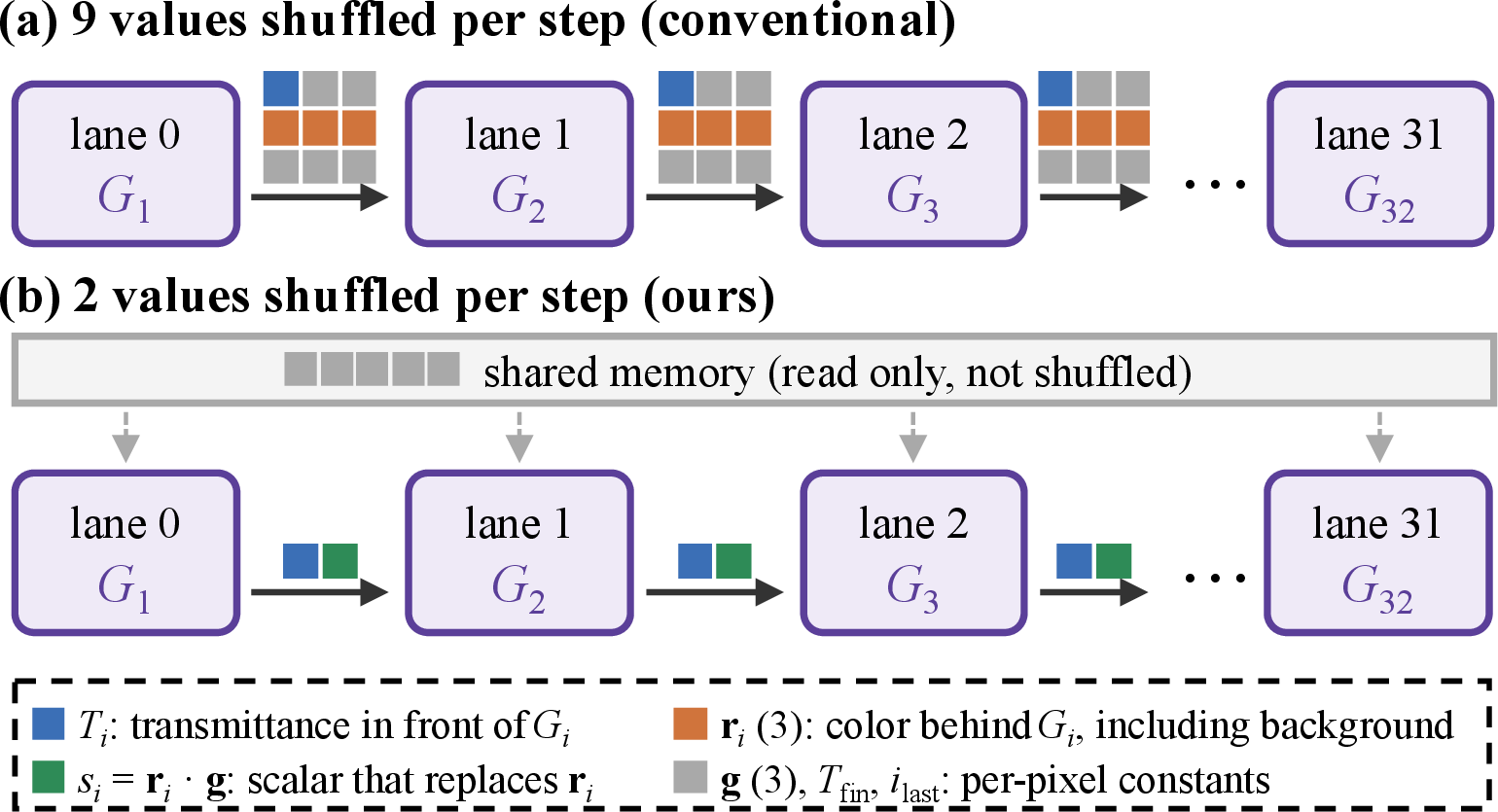}
  \caption{Splat-parallel backward pass; $\rightarrow$: warp shuffle to the next lane.
  SprintGS reads the per-pixel constants from shared memory and replaces $\mathbf{r}_i$ by the scalar $s_i$, so only 2 values are shuffled.}
  \label{fig:fc_shuffle}
\end{figure}

\noindent\textbf{Motivation.}
The backward pass of the rasterizer is the most expensive step (41\%, \cref{fig:breakdown}).
Fast rasterizers~\cite{Taming3DGS,FasterGS,FastGS,DashGaussian} parallelize it over Gaussians
and pass at least 9 per-pixel values between threads at every step (\cref{tab:shuffle});
F-C reduces them to 2.

\noindent\textbf{Splat-parallel backward pass.}
With $\mathbf{g}=\partial L/\partial\mathbf{C}$ for the pixel color $\mathbf{C}$ in \cref{eq:render}, the gradient with respect to $\alpha_i$ is
\begin{equation}
  \frac{\partial L}{\partial\alpha_i}
  = T_i(\mathbf{c}_i\!\cdot\!\mathbf{g})
  - \frac{\mathbf{r}_i\cdot\mathbf{g}}{1-\alpha_i},
  \qquad \mathbf{r}_i=\sum_{j>i}\mathbf{c}_j\,\alpha_j\,T_j+T_{\mathrm{fin}}\mathbf{b},
  \label{eq:dl_dalpha}
\end{equation}
where $\mathbf{r}_i$ is the color behind the $i$-th Gaussian, including the background.
Each thread of a warp (32 threads on NVIDIA GPUs) holds one Gaussian of the tile, and the pixels pass through the threads like a pipeline (\cref{fig:fc_shuffle}a): at each step, a thread updates the gradient of its Gaussian with its current pixel and hands the state of that pixel to the next thread by a \emph{warp shuffle}, a register exchange between threads.

\noindent\textbf{Reducing the state from 9 values to 2.}
Only $T_i$ and the scalar $s_i=\mathbf{r}_i\cdot\mathbf{g}$ change from thread to thread (\cref{fig:fc_shuffle}b).
The per-pixel constants $\mathbf{g}$, $T_{\mathrm{fin}}$, and the index $i_{\mathrm{last}}$ of the last contributing Gaussian (5 values)
are read from a small shared-memory window instead of being shuffled.
The color $\mathbf{r}_i$ (3 values) enters \cref{eq:dl_dalpha} only through $s_i$, so we pass $s_i$ instead.
It is initialized at the start of each bucket of 32 Gaussians from the color accumulated in front of the bucket
and is updated front to back with the already computed $\mathbf{c}_i\!\cdot\!\mathbf{g}$:
\begin{equation}
  s_i=s_{i-1}-\alpha_iT_i(\mathbf{c}_i\!\cdot\!\mathbf{g}),\qquad
  \frac{\partial L}{\partial\alpha_i}=T_i(\mathbf{c}_i\!\cdot\!\mathbf{g})-\frac{s_i}{1-\alpha_i}.
  \label{eq:fc_scalar}
\end{equation}
Both steps are algebraic rearrangements of \cref{eq:dl_dalpha}. The method applies to any splat-parallel rasterizer~\cite{Taming3DGS,FasterGS,DashGaussian}.

\begin{table}[t]
\centering
\small
\caption{Values passed between threads by warp shuffle at every step of the splat-parallel
backward pass (symbols as in \cref{sec:method_backward}; $\mathbf{r}_i$ and $\mathbf{g}$ have 3 values each; $\bar{b}$: precomputed background term).
$^\dagger$Two values for optional depth supervision, which is disabled in our experiments.}
\label{tab:shuffle}
\begin{tabular}{@{}lcl@{}}
\toprule
Method & \# & Shuffled values \\
\midrule
Taming-3DGS~\cite{Taming3DGS}    & 9  & $T_i,\ i_{\mathrm{last}},\ T_{\mathrm{fin}},\ \mathbf{r}_i(3),\ \mathbf{g}(3)$ \\
FastGS~\cite{FastGS}             & 9  & $T_i,\ i_{\mathrm{last}},\ T_{\mathrm{fin}},\ \mathbf{r}_i(3),\ \mathbf{g}(3)$ \\
DashGaussian~\cite{DashGaussian} & 9+2$^\dagger$ & $T_i,\ i_{\mathrm{last}},\ T_{\mathrm{fin}},\ \mathbf{r}_i(3),\ \mathbf{g}(3)$, depth \\
Faster-GS~\cite{FasterGS}        & 9  & $T_i,\ i_{\mathrm{last}},\ \bar{b},\ \mathbf{r}_i(3),\ \mathbf{g}(3)$ \\
\textbf{Ours}                    & \textbf{2} & $T_i,\ s_i$ \\
\bottomrule
\end{tabular}
\end{table}

%%% F-D
\subsection{Fused Loss with Forward-Pass Gradient (F-D)}
\label{sec:method_loss}

\noindent\textbf{Motivation.}
The loss takes 23\% of the training time of FastGS (\cref{fig:breakdown}). FastGS computes SSIM with the fused SSIM kernel of Taming-3DGS~\cite{Taming3DGS}, a CUDA kernel that computes SSIM and its local derivatives in one pass, but the L1 term, the weighting, and the mean are separate PyTorch operations, and the fused SSIM kernel still needs a backward kernel that convolves an upstream gradient map. F-D fuses the whole loss and moves its gradient computation into the forward pass.

\noindent\textbf{Gradient in the forward pass.}
By the chain rule, the image gradient is
\begin{equation}
  \frac{\partial L}{\partial C_q}
  = \sum_{p}\frac{\partial L}{\partial\ell_p}\,\frac{\partial\ell_p}{\partial C_q}
  = \frac{1}{M}\sum_{p}\frac{\partial\ell_p}{\partial C_q}.
  \label{eq:fd}
\end{equation}
Since $\partial L/\partial\ell_p=1/M$ is a constant, the right-hand side depends only on
the rendered and ground-truth images, which are available in the forward pass.
The autograd graph of FastGS does not exploit this:
it receives $\partial L/\partial\ell_p$ as an input of the backward pass and materializes it as a map.
We therefore compute both the scalar loss $L$ and the gradient $\partial L/\partial\mathbf{C}$
in the forward pass with three kernels:
\begin{itemize}[leftmargin=1.2em,itemsep=1pt,topsep=2pt,parsep=0pt]
  \item \textbf{Kernel 1} computes the local SSIM statistics and the derivatives of SSIM with respect to them,
  and reduces the L1 and SSIM terms to sums without writing a loss map.
  \item \textbf{Kernel 2} convolves these derivatives into the image gradient.
  \item \textbf{Kernel 3} forms $L$ from the sums with a single thread.
\end{itemize}
When $L$ is back-propagated without weighting, as in our pipelines, its upstream gradient is 1, and the backward pass simply returns the stored gradient without launching a kernel.
This removes the loss map, the mean reduction, the constant gradient map, and all backward kernels of the loss, including that of the fused SSIM.

%%% Table: Quantitative comparison  (SprintGS rows: tau_c = 0.10 on every dataset; Base and Big = 5-run means, 2026-11-09)
\begin{table*}[htbp]
\centering
\footnotesize
\setlength{\tabcolsep}{1.6pt}
\caption{Quantitative comparison on three benchmarks.
Time is the training time in seconds, and \#G the number of Gaussians in millions; the last two columns average all 13 scenes.
The best and second-best results are in \colorbox{tabfirstcolor}{red} and \colorbox{tabsecondcolor}{orange}.}
\label{tab:main_comparison}
\begin{tabular}{l ccccc ccccc ccccc cc}
\toprule
& \multicolumn{5}{c}{Mip-NeRF 360~\cite{MipNeRF360}} & \multicolumn{5}{c}{Deep Blending~\cite{DeepBlending}} & \multicolumn{5}{c}{Tanks \& Temples~\cite{TT}} & \multicolumn{2}{c}{All 13 scenes} \\
\cmidrule(lr){2-6}\cmidrule(lr){7-11}\cmidrule(lr){12-16}\cmidrule(lr){17-18}
Method & PSNR$\uparrow$ & SSIM$\uparrow$ & LPIPS$\downarrow$ & Time$\downarrow$ & \#G$\downarrow$
       & PSNR$\uparrow$ & SSIM$\uparrow$ & LPIPS$\downarrow$ & Time$\downarrow$ & \#G$\downarrow$
       & PSNR$\uparrow$ & SSIM$\uparrow$ & LPIPS$\downarrow$ & Time$\downarrow$ & \#G$\downarrow$
       & PSNR$\uparrow$ & Time$\downarrow$ \\
\midrule
3DGS~\cite{Vanilla3DGS}                  & 27.49 & 0.814 & 0.217 & 1487 & 2.72 & 29.76 & 0.903 & 0.241 & 1518 & 2.46 & 23.81 & 0.850 & 0.171 & 909 & 1.57 & 27.27 & 1403 \\
QuadBox~\cite{QuadBox}                   & 27.56 & 0.817 & 0.214 & 1348 & 2.78 & 29.84 & \tabsecond{0.908} & \tabsecond{0.237} & 1401 & 2.56 & 23.78 & 0.853 & 0.169 & 823 & 1.60 & 27.33 & 1275 \\
Mini-Splatting~\cite{mini-splating}      & 27.23 & 0.820 & 0.219 & 1230 & 0.50 & 29.95 & 0.907 & 0.243 & 1076 & 0.56 & 23.24 & 0.842 & 0.182 & 794 & 0.30 & 27.04 & 1139 \\
gsplat~\cite{gsplat}                     & 27.73 & \tabfirst{0.824} & \tabfirst{0.201} & 986 & 3.22 & 29.57 & 0.904 & 0.246 & 903 & 3.03 & 23.48 & 0.845 & 0.178 & 617 & 1.82 & 27.36 & 916 \\
Speedy-Splat~\cite{SpeedySplat}          & 26.83 & 0.783 & 0.291 & 924 & \tabfirst{0.31} & 29.56 & 0.899 & 0.272 & 834 & 0.25 & 23.39 & 0.816 & 0.242 & 516 & \tabfirst{0.18} & 26.72 & 847 \\
Turbo-GS~\cite{TurboGS_CVPR2026}        & 26.53 & 0.790 & 0.264 & 783 & 0.75 & 29.81 & 0.907 & 0.252 & 619 & 0.84 & 22.91 & 0.833 & 0.204 & 658 & 0.63 & 26.48 & 738 \\
Faster-GS~\cite{FasterGS}                & 27.53 & 0.816 & 0.212 & 375 & 2.69 & 29.74 & 0.906 & \tabfirst{0.236} & 318 & 2.57 & 23.55 & 0.850 & \tabsecond{0.168} & 247 & 1.54 & 27.26 & 347 \\
Taming-3DGS~\cite{Taming3DGS}            & 27.34 & 0.794 & 0.261 & 395 & 0.66 & 29.51 & 0.897 & 0.275 & 238 & 0.29 & 23.94 & 0.836 & 0.208 & 228 & 0.32 & 27.15 & 345 \\
DashGaussian~\cite{DashGaussian}         & 27.63 & 0.815 & 0.222 & 367 & 2.23 & 29.64 & 0.901 & 0.252 & 240 & 1.96 & 23.95 & 0.847 & 0.182 & 279 & 1.20 & 27.37 & 334 \\
ShorterSplatting~\cite{ShorterSplatting} & 27.23 & 0.809 & 0.225 & 211 & 3.31 & 29.26 & 0.897 & 0.249 & 171 & 2.80 & 23.40 & 0.843 & 0.172 & 203 & 1.83 & 26.95 & 203 \\
Mini-Splatting2~\cite{mini-splating2}    & 27.36 & 0.821 & 0.215 & 197 & 0.62 & 30.09 & \tabfirst{0.912} & 0.241 & 155 & 0.64 & 23.12 & 0.841 & 0.186 & 131 & 0.35 & 27.13 & 180 \\
SkipGS~\cite{SkipGS}                     & 27.37 & 0.794 & 0.265 & \tabsecond{175} & 0.37 & 30.00 & 0.901 & 0.272 & 131 & \tabsecond{0.22} & 23.98 & 0.836 & 0.213 & 129 & \tabsecond{0.24} & 27.26 & 161 \\
\midrule
FastGS-Base~\cite{FastGS}                & 27.44 & 0.795 & 0.263 & 186 & 0.37 & 29.95 & 0.901 & 0.270 & 143 & \tabsecond{0.22} & 24.09 & 0.839 & 0.210 & 138 & \tabsecond{0.24} & 27.31 & 172 \\
FastGS-Big~\cite{FastGS}                 & \tabsecond{27.83} & 0.817 & 0.219 & 317 & 1.02 & 30.09 & 0.907 & 0.243 & 210 & 0.65 & \tabfirst{24.45} & \tabsecond{0.854} & 0.176 & 203 & 0.54 & \tabsecond{27.65} & 283 \\
\midrule
SprintGS-Base (Ours)                     & 27.53 & 0.800 & 0.259 & \tabfirst{90} & \tabsecond{0.36} & \tabsecond{30.12} & 0.903 & 0.268 & \tabfirst{53} & \tabfirst{0.21} & 23.94 & 0.841 & 0.205 & \tabfirst{62} & \tabsecond{0.24} & 27.38 & \tabfirst{80} \\
SprintGS-Big (Ours)                      & \tabfirst{28.00} & \tabsecond{0.823} & \tabsecond{0.211} & 181 & 1.03 & \tabfirst{30.26} & \tabsecond{0.908} & 0.240 & \tabsecond{87} & 0.64 & \tabsecond{24.33} & \tabfirst{0.859} & \tabfirst{0.165} & \tabsecond{109} & 0.57 & \tabfirst{27.78} & \tabsecond{156} \\
\bottomrule
\end{tabular}
\end{table*}

%%% F-E
\subsection{Implementation Optimizations (F-E)}
\label{sec:method_impl}
Besides the rasterizer, each training iteration runs many small GPU operations,
such as parameter activations, optimizer updates, and updates of the densification statistics,
and the CPU waits for the GPU to read the loss.
These kernel-launch and waiting costs do not shrink with the model,
so their share of the training time grows as the model becomes compact (\cref{fig:breakdown}).
We adapt four implementation optimizations of Faster-GS~\cite{FasterGS} to the FastGS pipeline (F-E1--F-E4)
and extend two of them, F-E1 and F-E2 (details in the supplementary material).

\noindent\textbf{F-E1: Fused activations.}
As in Faster-GS, the activations of the parameters (exp, normalization, and sigmoid) and their derivatives
are computed inside the rasterizer kernels instead of as separate operations.
Going beyond Faster-GS, we further pass the gradients from the rasterizer to the optimizer without automatic differentiation, removing its intermediate tensors and gradient accumulation.

\noindent\textbf{F-E2: Single-launch Adam.}
Faster-GS computes the Adam update of each parameter type in one CUDA kernel,
instead of the several PyTorch operations of the standard Adam.
Building on this, we further update all parameter types with a single kernel call per optimizer.

\noindent\textbf{F-E3: Densification statistics in the backward kernel.}
Following Faster-GS, the backward kernel accumulates the densification statistics,
here adapted to those of FastGS (\eg, the accumulated view-space gradient norms and the maximum screen-space radius).
This removes the PyTorch operations that update them after each backward pass.

\noindent\textbf{F-E4: Asynchronous training loop.}
As in Faster-GS, the training loop does not wait for the GPU at every iteration: the loss is kept on the GPU and read only every 500 iterations.

%%% Section 5
\section{Experiments}
\label{sec:experiments}

%%% Section 5.1
\subsection{Experimental Setup}

\noindent\textbf{Datasets and metrics.} 
We evaluate on the 13 scenes used in 3DGS~\cite{Vanilla3DGS} from three real-world benchmarks:
all nine scenes of Mip-NeRF 360~\cite{MipNeRF360}, \textit{train} and \textit{truck} of Tanks \& Temples~\cite{TT}, and \textit{playroom} and \textit{drjohnson} of Deep Blending~\cite{DeepBlending}.
For a fair comparison, all methods share the camera poses and the initial sparse point clouds released by the authors of 3DGS~\cite{Vanilla3DGS}.
Following the common setting of 3DGS and its accelerated methods~\cite{Vanilla3DGS,Taming3DGS,SpeedySplat,DashGaussian,FasterGS}, we downsample Mip-NeRF 360 images to 1/4 (outdoor scenes) and 1/2 (indoor scenes) resolution
and hold out every eighth image of each scene for testing.
% Every eighth image is held out for testing.
We report PSNR, SSIM~\cite{SSIM}, LPIPS (VGG)~\cite{LPIPS}, the training time (Time) in seconds, and the number of Gaussians in the final model (\#G) in millions.
For all methods, Time is the wall-clock time of the training loop, measured with the same host-side timer and CUDA synchronization at both ends; it excludes data loading, evaluation, logging, and saving.

\noindent\textbf{Implementation details.}
SprintGS-Base and SprintGS-Big, which prioritize training speed and rendering quality, respectively,
build on the \textit{base} and \textit{big} configurations of FastGS~\cite{FastGS}
and inherit all their hyperparameters, including the per-scene ones, with $T=30{,}000$ iterations
(\textit{big} densifies more often with lower gradient thresholds, yielding two to three times more Gaussians).
For both, we use $\tau_c=0.1$, $\theta_o=0.1$, $q=0.5$, $K=500$, and $\rho=0.4$; their sensitivity is analyzed in the supplementary material.
All results except those on DL3DV-10K~\cite{DL3DV-10K} (\cref{tab:dl3dv}, one run per scene) are averaged over five runs on a single NVIDIA RTX 3090 Ti; the speed-up holds on other GPUs (see the supplementary material).

\begin{figure*}[t]
\centering
\includegraphics[width=\linewidth]{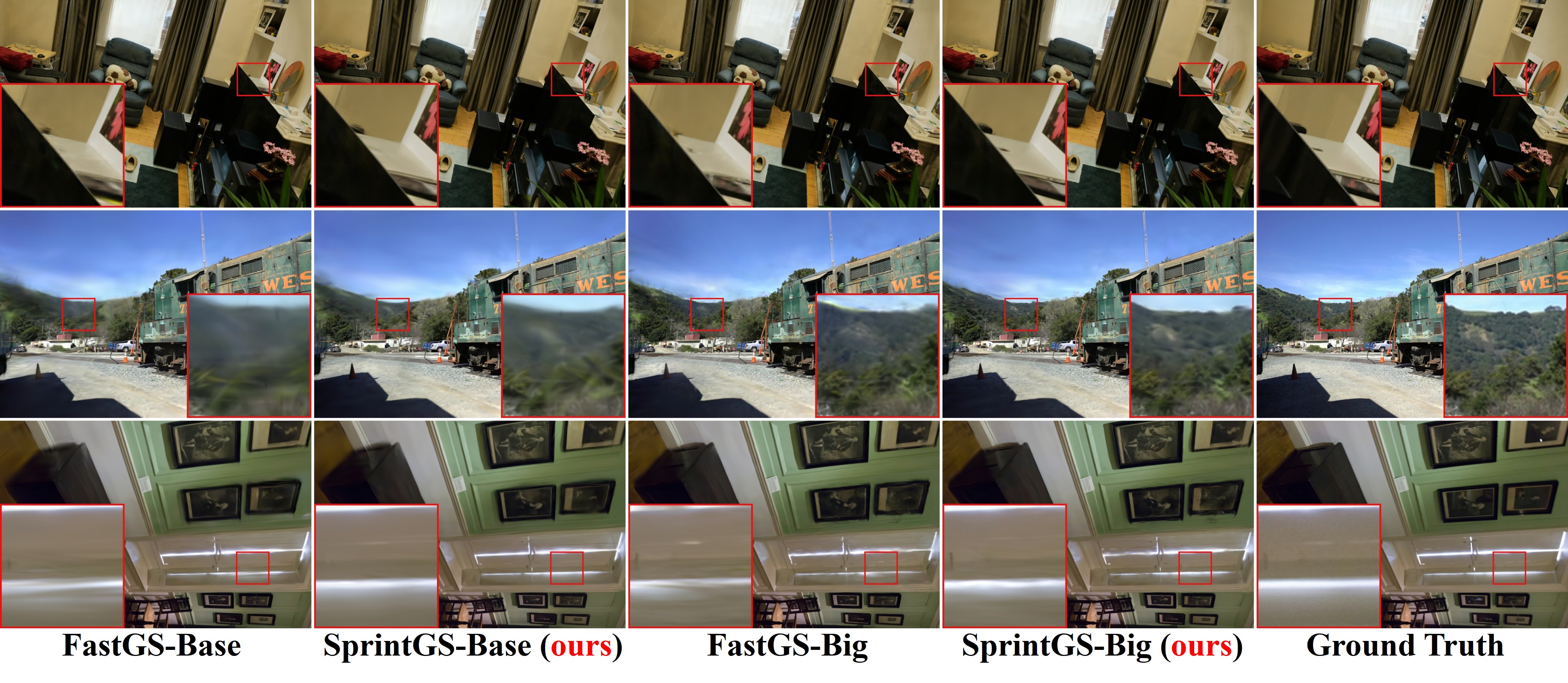}
\caption{Qualitative comparison on test views (zoomed-in regions in the insets).
From top to bottom: \textit{room} from Mip-NeRF 360~\cite{MipNeRF360}, \textit{train} from Tanks \& Temples~\cite{TT}, and \textit{drjohnson} from Deep Blending~\cite{DeepBlending}.}
  \label{fig:render_results}
\end{figure*}

\begin{table}[t]
\centering
\footnotesize
\setlength{\tabcolsep}{3pt}
\caption{Results on the 1K subset of DL3DV-10K~\cite{DL3DV-10K} (average over 1,000 scenes, one run each). The better results are in \textbf{bold}.}
\label{tab:dl3dv}
\begin{tabular}{@{}lccccc@{}}
\toprule
Method & PSNR$\uparrow$ & SSIM$\uparrow$ & LPIPS$\downarrow$ & Time$\downarrow$ & \#G$\downarrow$ \\
\midrule
FastGS-Base~\cite{FastGS} & 28.42 & 0.881 & 0.184 & 141.8 & \textbf{0.14} \\
SprintGS-Base (Ours)      & \textbf{28.51} & \textbf{0.883} & \textbf{0.182} & \textbf{63.3} & \textbf{0.14} \\
\bottomrule
\end{tabular}
\end{table}

\begin{table}[t]
\centering
\footnotesize
\setlength{\tabcolsep}{3pt}
\caption{Ablation study removing each component of SprintGS-Base in turn (averages over all 13 scenes). $\Delta$Time is relative to the full model (SprintGS-Base).}
\label{tab:ablation}
\begin{tabular}{l cccccc}
\toprule
Configuration & PSNR$\uparrow$ & SSIM$\uparrow$ & LPIPS$\downarrow$ & Time$\downarrow$ & $\Delta$Time & \#G$\downarrow$ \\
\midrule
SprintGS-Base           & 27.38 & 0.822 & 0.252 & 80.0  & --     & 0.318 \\
\midrule
w/o F-A                 & 27.39 & 0.822 & 0.251 & 89.3  & +12\%  & 0.317 \\
w/o F-B                 & 27.29 & 0.819 & 0.256 & 81.8  & +2\%   & 0.337 \\
w/o F-C                 & 27.37 & 0.822 & 0.252 & 98.2  & +23\%  & 0.318 \\
w/o F-D                 & 27.37 & 0.822 & 0.252 & 104.5 & +31\%  & 0.318 \\
w/o F-E1                & 27.38 & 0.822 & 0.252 & 91.1  & +14\%  & 0.318 \\
w/o F-E2                & 27.37 & 0.822 & 0.252 & 84.3  & +5\%   & 0.318 \\
w/o F-E3                & 27.37 & 0.822 & 0.252 & 86.6  & +8\%   & 0.318 \\
w/o F-E4                & 27.38 & 0.822 & 0.252 & 84.7  & +6\%   & 0.318 \\
\bottomrule
\end{tabular}
\end{table}

\subsection{Comparison with Fast Training Methods}

\noindent\textbf{Baselines.}
We compare SprintGS with 3DGS~\cite{Vanilla3DGS} and 12 fast training methods~\cite{gsplat,mini-splating2,mini-splating,TurboGS_CVPR2026,FasterGS,Taming3DGS,DashGaussian,ShorterSplatting,SkipGS,FastGS,QuadBox,SpeedySplat}, all of which are run with their official code and the settings recommended by the authors.
For Taming-3DGS~\cite{Taming3DGS} and DashGaussian~\cite{DashGaussian}, we use the budget mode and the sparse-Adam configuration, respectively, which are the fastest settings reported in their papers.
For SkipGS~\cite{SkipGS}, a plug-in that skips redundant backward passes, we use its released integration with FastGS~\cite{FastGS}, the fastest backbone.
All methods are trained for 30{,}000 iterations as in SprintGS, except Mini-Splatting2~\cite{mini-splating2}, which uses 18,000 iterations as in its official setting.

\noindent\textbf{Results on 13 scenes.}
As shown in \cref{tab:main_comparison}, SprintGS-Base trains the fastest on all three datasets.
Averaged over all 13 scenes, it accelerates FastGS-Base~\cite{FastGS} by $2.15\times$ (172\,s $\to$ 80\,s) at comparable rendering quality (27.38\,dB vs.\ 27.31\,dB PSNR).
SprintGS-Big trains $1.81\times$ faster than FastGS-Big (283\,s $\to$ 156\,s), and even faster than FastGS-Base,
while achieving the highest average PSNR (27.78\,dB).
In \cref{fig:render_results}, FastGS blurs the laptop outline in \textit{room} and the distant vegetation on the hillside in \textit{train}, and smears the thin highlights on the door frame in \textit{drjohnson}, whereas SprintGS reproduces these fine structures more faithfully.

%\noindent\textbf{Consistency across GPUs.}
%Besides the RTX 3090 Ti (Ampere) used above, we train FastGS-Base and %SprintGS-Base
%on a Quadro RTX 6000 (Turing) and an RTX 5090 (Blackwell).
%SprintGS-Base is more than $2\times$ faster on all three GPUs
%see the supplementary material).

\noindent\textbf{Results on diverse scenes.}
We further evaluate on the 1K subset (1,000 scenes) of DL3DV-10K~\cite{DL3DV-10K}, using the released $960\times540$ images, camera poses, and sparse point clouds, and holding out every eighth image as above.
As shown in \cref{tab:dl3dv}, SprintGS-Base trains more than $2\times$ faster than FastGS-Base~\cite{FastGS} with slightly better quality, confirming that the speed-up holds beyond the 13 scenes.

\subsection{Ablation Study}
\label{sec:ablation}

\Cref{tab:ablation} removes each component of SprintGS-Base in turn, replacing it with the corresponding implementation of FastGS~\cite{FastGS}.
Every component shortens training, and F-C, F-D, and F-E barely affect the quality.
F-A slightly lowers it, as its coarse footprint drops small contributions, whereas F-B improves it with fewer Gaussians.

\subsection{SprintGS with Other Fast Methods}
\label{sec:other_baselines}

\begin{table}[t]
\centering
\footnotesize
\setlength{\tabcolsep}{3pt}
\caption{SprintGS combined with other fast training methods (averages over all 13 scenes). The better result is in \textbf{bold}.}
\label{tab:other_baselines}
\begin{tabular}{@{}lccccc@{}}
\toprule
Method & PSNR$\uparrow$ & SSIM$\uparrow$ & LPIPS$\downarrow$ & Time$\downarrow$ & \#G$\downarrow$ \\
\midrule
Taming-3DGS~\cite{Taming3DGS}       & 27.15 & 0.816 & 0.255 & 345 & 0.55 \\
+SprintGS (Ours)                 & \textbf{27.18} & \textbf{0.823} & \textbf{0.245} & \textbf{125} & \textbf{0.53} \\
\midrule
DashGaussian~\cite{DashGaussian} & 27.37 & 0.833 & 0.221 & 334 & 2.03 \\
+SprintGS (Ours)                 & \textbf{27.38}& \textbf{0.839} & \textbf{0.211} & \textbf{193} & \textbf{1.76} \\
\midrule
SkipGS~\cite{SkipGS}             & 27.26 & 0.817 & 0.258 & 161 & 0.33 \\
+SprintGS (Ours)                 & \textbf{27.29} & \textbf{0.821} & \textbf{0.254} & \textbf{76} & \textbf{0.32} \\
\bottomrule
\end{tabular}
\end{table}

Most components of SprintGS do not depend on FastGS~\cite{FastGS}.
To show this, we combine SprintGS with Taming-3DGS~\cite{Taming3DGS} (budget-controlled densification),
DashGaussian~\cite{DashGaussian} (resolution scheduling), and SkipGS~\cite{SkipGS} (skipping backward passes),
keeping their strategies.
SprintGS accelerates Taming-3DGS, DashGaussian, and SkipGS by $2.76\times$, $1.73\times$, and $2.12\times$, respectively, while preserving quality with fewer Gaussians (\cref{tab:other_baselines}).

\begin{table}[t]
\centering
\footnotesize
\setlength{\tabcolsep}{3pt}
\caption{Dynamic scene reconstruction on Neu3D~\cite{Neu3D} (average over six scenes). The best results are in \textbf{bold}.}
\label{tab:dynamic}
\begin{tabular}{@{}lccccc@{}}
\toprule
Method & PSNR$\uparrow$ & SSIM$\uparrow$ & LPIPS$\downarrow$ & Time$\downarrow$ & \#G$\downarrow$ \\
\midrule
Deformable-3DGS~\cite{Deformable3DGS} & 29.00 & 0.931 & 0.156 & 2609 & 0.24  \\
+ FastGS~\cite{FastGS}                & 30.83 & 0.941 & 0.147 & 804 & \textbf{0.08} \\
+ SprintGS (Ours)                     & \textbf{31.36} & \textbf{0.944} & \textbf{0.146} & \textbf{637} & \textbf{0.08}  \\
\bottomrule
\end{tabular}
\end{table}

\begin{figure*}[t]
  \centering
  \includegraphics[width=0.85\textwidth]{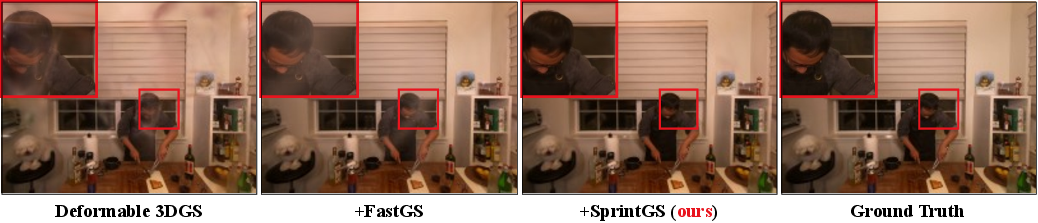}
  \caption{Qualitative results of dynamic scene reconstruction on the Neu3D dataset~\cite{Neu3D} (\textit{cut\_roasted\_beef}).}
  \label{fig:other_tasks}
\end{figure*}

\subsection{Extension to Other Reconstruction Tasks}
\label{sec:extension}

FastGS~\cite{FastGS} releases dynamic, sparse-view, and surface reconstruction methods that use FastGS as the backbone; we replace FastGS in them with SprintGS.

\noindent\textbf{Dynamic reconstruction.}
We apply SprintGS to Deformable-3DGS~\cite{Deformable3DGS} and evaluate on the six scenes of Neu3D~\cite{Neu3D}.
As shown in \cref{tab:dynamic,fig:other_tasks}, SprintGS trains $1.26\times$ faster than Deformable-3DGS+FastGS and improves PSNR by 0.53\,dB, with similar numbers of Gaussians.
In \cref{fig:other_tasks}, SprintGS reduces the haze around the moving person relative to both baselines.
The speed-up is smaller than on static scenes because every iteration also runs the deformation network of Deformable-3DGS,
a multilayer perceptron (MLP) whose cost SprintGS leaves unchanged.

\begin{table}[t]
\centering
\footnotesize
\setlength{\tabcolsep}{3pt}
\caption{Sparse-view reconstruction on LLFF~\cite{LLFF}.
The best results for each number of views are in \textbf{bold}.}
\label{tab:fewshot}
\begin{tabular}{@{}clccccc@{}}
\toprule
Views & Method & PSNR$\uparrow$ & SSIM$\uparrow$ & LPIPS$\downarrow$ & Time$\downarrow$ & \#G$\downarrow$ \\
\midrule
\multirow{3}{*}{3} & DropGaussian~\cite{DropGaussian} & 20.46 & 0.709 & \textbf{0.201} & 67 & 0.076 \\
 & +FastGS~\cite{FastGS} & 20.51 & 0.708 & 0.218 & 28 & 0.034 \\
 & +SprintGS (Ours) & \textbf{20.67} & \textbf{0.710} & 0.218 & \textbf{16} & \textbf{0.028} \\
\midrule
\multirow{3}{*}{6} & DropGaussian~\cite{DropGaussian} & 24.64 & \textbf{0.836} & \textbf{0.117} & 81 & 0.184 \\
 & +FastGS~\cite{FastGS} & 24.69 & 0.833 & 0.131 & 36 & 0.096 \\
 & +SprintGS (Ours) & \textbf{24.70} & 0.835 & 0.127 & \textbf{17} & \textbf{0.086} \\
\midrule
\multirow{3}{*}{9} & DropGaussian~\cite{DropGaussian} & 26.09 & 0.873 & \textbf{0.089} & 118 & 0.419 \\
 & +FastGS~\cite{FastGS} & 26.12 & 0.873 & 0.095 & 56 & 0.223 \\
 & +SprintGS (Ours) & \textbf{26.27} & \textbf{0.876} & \textbf{0.089} & \textbf{29} & \textbf{0.190} \\
\bottomrule
\end{tabular}
\end{table}

\noindent\textbf{Sparse-view reconstruction.}
We apply SprintGS to DropGaussian~\cite{DropGaussian}, a sparse-view 3DGS method, and evaluate on the eight scenes of the Local Light Field Fusion (LLFF) dataset~\cite{LLFF} with 3, 6, and 9 training views, following the protocol of DropGaussian.
As shown in \cref{tab:fewshot}, SprintGS trains $1.8$--$2.1\times$ faster than FastGS with the fewest Gaussians and the best PSNR in all settings.

\begin{table}[t]
\centering
\footnotesize
\setlength{\tabcolsep}{3pt}
\caption{Surface reconstruction on Tanks \& Temples~\cite{TT} (average over four scenes). The best results are in \textbf{bold}.}
\label{tab:surface}
\begin{tabular}{@{}lcccccc@{}}
\toprule
Method & PSNR$\uparrow$ & SSIM$\uparrow$ & LPIPS$\downarrow$ & Time$\downarrow$ & \#G$\downarrow$ & F-score$\uparrow$ \\
\midrule
PGSR~\cite{PGSR}      & 23.15 & \textbf{0.845} & \textbf{0.155} & 3263 & 1.73 & 0.557 \\
+FastGS~\cite{FastGS} & \textbf{23.38} & 0.836 & 0.194 & 1825 & 0.44 & 0.545 \\
+SprintGS (Ours)      & 23.26 & 0.837 & 0.192 & \textbf{618} & \textbf{0.43} & \textbf{0.560} \\
\bottomrule
\end{tabular}
\end{table}

\noindent\textbf{Surface reconstruction.}
We apply SprintGS to PGSR~\cite{PGSR}, a planar-based Gaussian splatting method for surface reconstruction, and evaluate the extracted meshes on four Tanks \& Temples~\cite{TT} scenes used in the official code of FastGS~\cite{FastGS} (\textit{truck}, \textit{caterpillar}, \textit{ignatius}, and \textit{meetingroom}).
To evaluate the geometric accuracy, we report the F-score against the ground-truth point cloud, computed with the official evaluation toolkit~\cite{TT}.
As shown in \cref{tab:surface}, SprintGS trains $3.0\times$ faster than FastGS. Its F-score is on par with PGSR (0.560 vs.\ 0.557), whereas FastGS falls below it (0.545).

%%% Section 6
\section{Limitations}
\label{sec:limitation}

\noindent\textbf{Training schedule.}
Our culling schedule (\cref{sec:method_scheduled-tile-culling}) is tied to $T$: checkpoints before $0.9T$ are trained with approximate footprints and may degrade under exact rendering. We therefore set $T$ to the intended checkpoint budget.

\noindent\textbf{Untouched costs.}
The speed-up is smaller when costs that SprintGS does not touch dominate, such as the deformation network in dynamic reconstruction.

\noindent\textbf{Loss form.}
F-D assumes that the loss is a plain mean of per-pixel terms; per-pixel weights or other loss terms require adapting its forward-pass gradient.

%%% Section 7
\section{Conclusion}
\label{sec:conclusion}
We presented SprintGS, a fast training framework built on FastGS~\cite{FastGS}
that reduces the redundant work in rasterization and the cost of the loss computation while keeping the model compact.
It trains $2.15\times$ faster than FastGS at comparable quality and $1.81\times$ faster in its high-quality setting.
In future work, we will evaluate SprintGS on large-scale scenes, where more Gaussians and tiles may make the redundant work in rasterization even more costly.

{
    \small
    \bibliographystyle{ieeenat_fullname}
    \bibliography{main}
}

%%%%%%%%%%%%%%%%%%%%%%%%%%%%%%%%%%%%%%%%%%%%%%%%%%%%%%%%%%%%%%%%%%%%%%%%%%%%%%%%
%%%%%%%%% SUPPLEMENTARY MATERIAL
%%%%%%%%%%%%%%%%%%%%%%%%%%%%%%%%%%%%%%%%%%%%%%%%%%%%%%%%%%%%%%%%%%%%%%%%%%%%%%%%
\appendix
\counterwithin{table}{section}
\counterwithin{figure}{section}
\counterwithin{equation}{section}
\counterwithin{algorithm}{section}

% \twocolumn starts a new page and prints a full-width heading
\twocolumn[{%
  \centering
  \Large\bfseries Supplementary Material\par
  \vspace{1.5em}
}]

\section*{Overview}
This supplementary material provides additional results and details that support the main text.
\begin{itemize}[leftmargin=1.2em,itemsep=1pt,topsep=2pt,parsep=0pt]
  \item \Cref{sec:supp_bench} reports additional results: the run-to-run variation, the training time on other GPUs, the rendering speed, the GPU memory consumption, the training-time breakdown, a cumulative ablation, and the rendering results on the DL3DV-10K dataset~\cite{DL3DV-10K}.
  \item \Cref{sec:supp_other_baselines} details the combinations with other fast methods (Sec.~5.4).
  \item \Cref{sec:supp_extension} details the extensions to dynamic, sparse-view, and surface reconstruction (Sec.~5.5).
  \item \Cref{sec:supp_components} details each component of SprintGS (Secs.~4.1--4.5),
  ablates its design choices, and analyzes its sensitivity to the hyperparameters.
\end{itemize}
Sections, tables, figures, and equations numbered without a letter refer to the main text.
Unless stated otherwise, all results are averages over five runs on an NVIDIA RTX 3090 Ti with the settings of the main text.

% ===================================================
% A. Additional Results
\section{Additional Experimental Results}
\label{sec:supp_bench}
This section supplements the experiments in Sec.~5
with results that are omitted from the main text due to space limitations.

%% Section A.1
\subsection{Run-to-Run Variation}
\label{sec:supp_variance}
As in 3DGS~\cite{Vanilla3DGS} and its variants, including FastGS~\cite{FastGS}, the SprintGS rasterizer accumulates gradients with floating-point atomic additions, whose order is not deterministic, so repeated runs with the same settings give slightly different results.
Except for DL3DV-10K (one run per scene), the evaluation results in the main text are averaged over five runs.
Here, we report how much these values vary between the runs. \Cref{tab:supp_variance} gives the mean and the standard deviation of the 13-scene average of SprintGS-Base over the five runs, trained with the settings of the main text.
The quality metrics and the number of Gaussians vary little between runs.
The standard deviation of the training time is 0.3\,s, which is much smaller than the speed-up over the conventional methods in Tab.~2, so the speed-up is not an effect of run-to-run variation.

%% Section A.2
\subsection{Training Time on Different GPUs}
\label{sec:supp_other_gpu}
To confirm that the speed-up of SprintGS does not depend on a particular GPU, we train FastGS-Base~\cite{FastGS} and SprintGS-Base on three GPUs of different generations: an NVIDIA RTX 5090 (Blackwell), the NVIDIA RTX 3090 Ti (Ampere) used in the main text, and an NVIDIA Quadro RTX 6000 (Turing). All machines use the same CPU (Intel Core i7-14700K), and all other settings follow the main text.
As shown in \cref{tab:supp_other_gpu}, SprintGS-Base trains $2.02$--$2.33\times$ faster than FastGS-Base on all three GPUs.

%%% Table: run-to-run variation
\begin{table}[t]
  \centering
  \footnotesize
  \setlength{\tabcolsep}{4pt}
  \caption{Variation of SprintGS-Base over five runs (average over all 13 scenes):
  the mean $\pm$ the standard deviation over the five runs.}
  \label{tab:supp_variance}
  \begin{tabular}{lc}
    \toprule
    & SprintGS-Base \\
    \midrule
    PSNR$\uparrow$       & 27.38 $\pm$ 0.01    \\
    SSIM$\uparrow$       & 0.8219 $\pm$ 0.0001 \\
    LPIPS$\downarrow$    & 0.2523 $\pm$ 0.0002 \\
    \#G (M)$\downarrow$  & 0.3178 $\pm$ 0.0003 \\
    Time (s)$\downarrow$ & 80.0 $\pm$ 0.3      \\
    \bottomrule
  \end{tabular}
\end{table}

%%% Table: Comparison across different GPUs
\begin{table}[t]
  \centering
  \footnotesize
  \setlength{\tabcolsep}{3pt}
  \caption{Training time (s) on GPUs of different generations, averaged over all 13 scenes and five runs.
  The architecture of each GPU is shown in parentheses.}
  \label{tab:supp_other_gpu}
  \begin{tabular}{lccc}
    \toprule
    & RTX 5090 & RTX 3090 Ti & Quadro RTX 6000 \\
    & (Blackwell) & (Ampere) & (Turing) \\
    \midrule
    FastGS-Base          & 105 & 172 & 249 \\
    SprintGS-Base (Ours) & \textbf{45} & \textbf{80} & \textbf{123} \\
    \midrule
    Speed-up             & $2.33\times$ & $2.15\times$ & $2.02\times$ \\
    \bottomrule
  \end{tabular}
\end{table}

\subsection{Rendering Speed}
\label{sec:supp_fps}

We measure the rendering speed of the final models of all methods in Tab.~2 of the main paper with two renderers.
\begin{itemize}[leftmargin=1.2em,itemsep=1pt,topsep=2pt,parsep=0pt]
  \item \textbf{3DGS renderer~\cite{Vanilla3DGS}.} The final Gaussians of every method are rendered with the official renderer of the original 3DGS~\cite{Vanilla3DGS}. Since all methods share this renderer, the speed depends only on the 3DGS model.
  \item \textbf{Own renderer.} Each model is rendered with the renderer of its own method, loaded and called as in the official rendering script of the method.
\end{itemize}
We render the test views at the evaluation resolution. After 10 warm-up frames, we render all test views five times, synchronize the GPU before and after each pass, and report the number of frames per second (FPS), which includes only the rendering call (no metric computation or image I/O). \Cref{tab:supp_fps} reports the averages over all 13 scenes.

%%% Table: rendering speed
\begin{table}[t]
  \centering
  \footnotesize
  \setlength{\tabcolsep}{4pt}
  \caption{Rendering speed (FPS, averages over all 13 scenes) with the renderer of 3DGS~\cite{Vanilla3DGS} for all methods (3DGS) and with the renderer of each method (Own). The best and second-best results are in \colorbox{tabfirstcolor}{red} and \colorbox{tabsecondcolor}{orange}.}
  \label{tab:supp_fps}
  \begin{tabular}{lccc}
    \toprule
    & \multicolumn{2}{c}{Renderer} & \\
    \cmidrule(lr){2-3}
    Method & 3DGS$\uparrow$ & Own$\uparrow$ & \#G$\downarrow$ \\
    \midrule
    3DGS~\cite{Vanilla3DGS}                  & 131 & 131 & 2.50 \\
    QuadBox~\cite{QuadBox}                   & 128 & 186 & 2.56 \\
    Mini-Splatting~\cite{mini-splating}      & \tabsecond{362} & 520 & 0.48 \\
    gsplat~\cite{gsplat}                     & 130 & 327 & 2.98 \\
    Speedy-Splat~\cite{SpeedySplat}          & \tabfirst{410} & 605 & \tabfirst{0.28} \\
    Turbo-GS~\cite{TurboGS_CVPR2026}         & 222 & 100 & 0.75 \\
    Faster-GS~\cite{FasterGS}                & 129 & \tabsecond{677} & 2.49 \\
    Taming-3DGS~\cite{Taming3DGS}            & 214 & 267 & 0.55 \\
    DashGaussian~\cite{DashGaussian}         & 109 & 158 & 2.03 \\
    ShorterSplatting~\cite{ShorterSplatting} & 192 & 297 & 3.00 \\
    Mini-Splatting2~\cite{mini-splating2}    & 334 & 466 & 0.58 \\
    SkipGS~\cite{SkipGS}                     & 329 & 580 & 0.33 \\
    \midrule
    FastGS-Base~\cite{FastGS}                & 322 & 571 & 0.33 \\
    FastGS-Big~\cite{FastGS}                 & 261 & 477 & 0.89 \\
    \midrule
    SprintGS-Base (Ours)                     & 340 & \tabfirst{955} & \tabsecond{0.32} \\
    SprintGS-Big (Ours)                      & 262 & 657 & 0.90 \\
    \bottomrule
  \end{tabular}
\end{table}

\noindent\textbf{Results with the 3DGS renderer.}
With the same renderer, the speed generally decreases with the number of Gaussians.
However, this is only a trend: Mini-Splatting~\cite{mini-splating} renders faster than our SprintGS-Base and FastGS-Base~\cite{FastGS} with more Gaussians.
The cost of the renderer also depends on the size of each Gaussian on the screen
and on how many Gaussians overlap each pixel,
so we presume that the Gaussians of Mini-Splatting are smaller on the screen or overlap less.
SprintGS-Base is the third fastest after Speedy-Splat~\cite{SpeedySplat} and Mini-Splatting~\cite{mini-splating},
and slightly faster than its direct baseline FastGS-Base~\cite{FastGS}.

\noindent\textbf{Results with the own renderers.}
Most methods render faster with their own renderer than with that of 3DGS~\cite{Vanilla3DGS},
and SprintGS-Base reaches the highest speed (955\,FPS).

\noindent\textbf{Renderer of SprintGS.}
SprintGS targets the training time; we report the rendering speed for reference.
At these speeds, a frame takes only 1--2\,ms, so the fixed cost of the PyTorch operations around the rasterizer is a noticeable part of it.
The renderer of SprintGS is that of FastGS except for two changes inherited from training that reduce this cost:
(i) the activations of the scale, rotation, and opacity are applied inside the rasterizer (F-E1), and
(ii) the screen-space point tensor, which FastGS creates in every call only to collect the gradients for densification, is not created, since F-E3 accumulates these statistics in the backward kernel.
The speed-up does not come from a coarser tile assignment: SprintGS renders with the exact footprint ($\tau=\epsilon$, Sec.~4.1 of the main paper).

%%%%%%%%%% Section A.5 %%%%%%%%%%
\subsection{GPU Memory Consumption}
\label{sec:supp_mem}

%%% Table: peak training memory
\begin{table}[t]
  \centering
  \footnotesize
  \setlength{\tabcolsep}{4pt}
  \caption{Peak GPU memory of the training (GiB),
  averaged over the scenes of Mip-NeRF 360 (M360), Tanks\&Temples (T\&T), and Deep Blending (DB),
  and over all 13 scenes (All).
  The best and second-best results are in \colorbox{tabfirstcolor}{red} and \colorbox{tabsecondcolor}{orange}.}
  \label{tab:supp_mem}
  \begin{tabular}{lcccc}
    \toprule
    Method & M360$\downarrow$ & T\&T$\downarrow$ & DB$\downarrow$ & All$\downarrow$ \\
    \midrule
    3DGS~\cite{Vanilla3DGS}                  & 8.80  & 4.73  & 8.11  & 8.06 \\
    QuadBox~\cite{QuadBox}                   & 8.22  & 4.37  & 7.58  & 7.53 \\
    Mini-Splatting~\cite{mini-splating}      & 6.38  & 4.64  & 6.21  & 6.09 \\
    gsplat~\cite{gsplat}                     & 4.86  & 2.79  & 4.55  & 4.49 \\
    Speedy-Splat~\cite{SpeedySplat}          & 8.34  & 4.48  & 6.82  & 7.51 \\
    Turbo-GS~\cite{TurboGS_CVPR2026}         & 6.42  & 3.98  & 6.15  & 6.01 \\
    Faster-GS~\cite{FasterGS}                & 5.94  & 3.25  & 5.77  & 5.50 \\
    Taming-3DGS~\cite{Taming3DGS}            & 4.75  & 2.37  & 3.91  & 4.25 \\
    DashGaussian~\cite{DashGaussian}         & 8.40  & 4.49  & 7.83  & 7.71 \\
    ShorterSplatting~\cite{ShorterSplatting} & 12.36 & 6.27  & 11.15 & 11.23 \\
    Mini-Splatting2~\cite{mini-splating2}    & 8.24  & 5.49  & 8.19  & 7.81 \\
    SkipGS~\cite{SkipGS}                     & 4.51  & 2.44  & 4.03  & 4.11 \\
    \midrule
    FastGS-Base~\cite{FastGS}                & \tabsecond{4.26} & \tabsecond{2.27} & \tabsecond{3.78} & \tabsecond{3.88} \\
    FastGS-Big~\cite{FastGS}                 & 5.49  & 2.83  & 4.66  & 4.95 \\
    \midrule
    SprintGS-Base (Ours)                     & \tabfirst{4.08} & \tabfirst{2.10} & \tabfirst{3.60} & \tabfirst{3.70} \\
    SprintGS-Big (Ours)                      & 5.24  & 2.68  & 4.32  & 4.71 \\
    \bottomrule
  \end{tabular}
\end{table}

We measure the peak GPU memory during the training, excluding the evaluation, for all methods in Table~2 of the main paper.

\noindent\textbf{Results.}
\Cref{tab:supp_mem} reports the averages over the scenes of each dataset and over all 13 scenes. SprintGS-Base uses the least memory on every dataset,
and both SprintGS configurations use slightly less memory than their FastGS~\cite{FastGS} counterparts.
The peak memory depends on the largest model during the training rather than on the final model. We presume that SprintGS keeps slightly fewer Gaussians than FastGS during the training, since F-B removes converged transparent Gaussians, and that this contributes to its smaller memory. In contrast, Speedy-Splat~\cite{SpeedySplat} ends with the fewest Gaussians (Tab.~\ref{tab:supp_fps}) but uses about twice as much memory as SprintGS-Base, because by design it keeps many more Gaussians in the first half of the training before pruning most of them.

\subsection{Training-Time Breakdown of SprintGS}
\label{sec:supp_breakdown}

\begin{figure}[t]
  \centering
  \includegraphics[width=\linewidth]{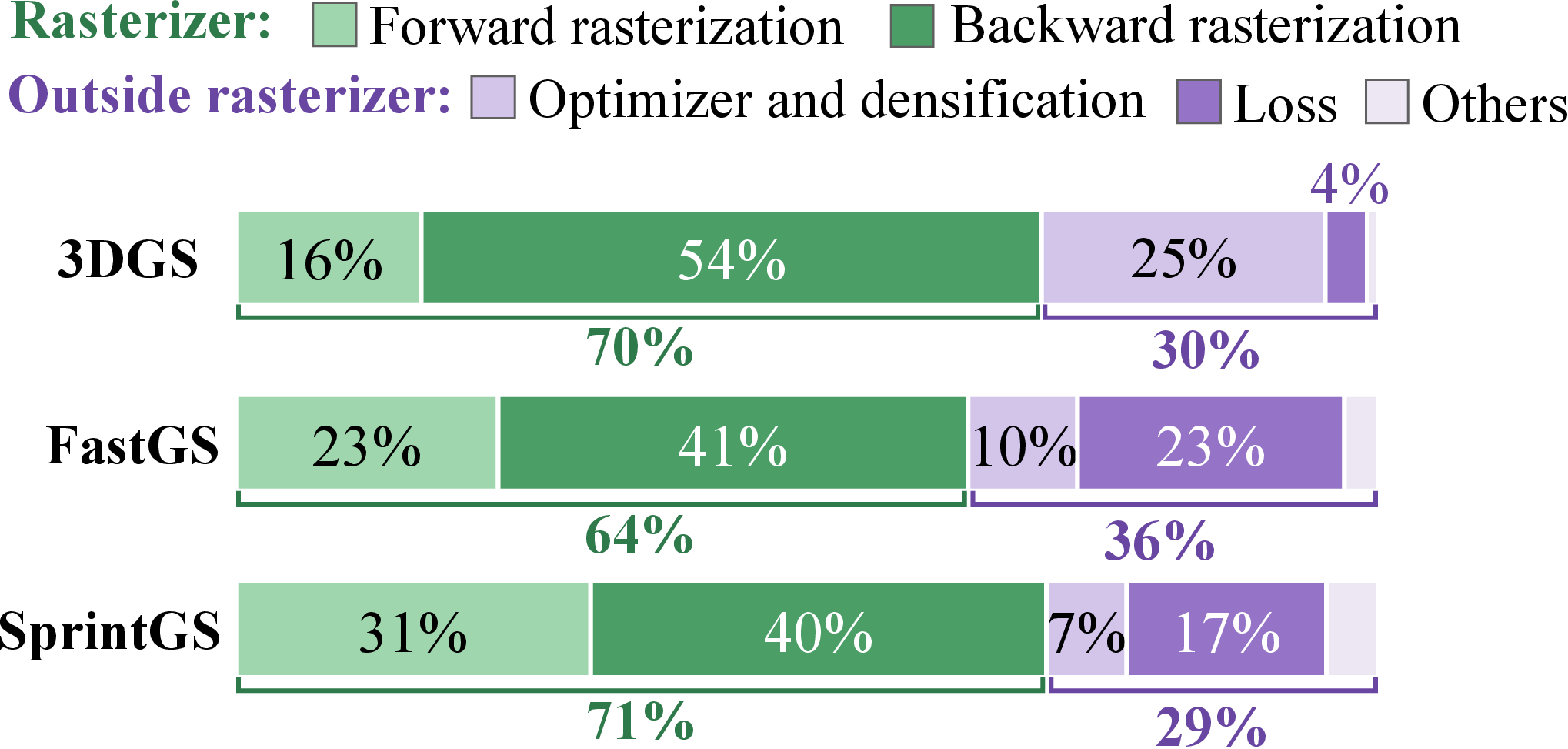}
  \caption{Training-time breakdown of 3DGS~\cite{Vanilla3DGS}, FastGS-Base~\cite{FastGS}, and SprintGS-Base
  over the 13 scenes, measured with a synchronization after each step.
  Each bar is normalized to the total training time of the method.
  FastGS and SprintGS in the figure denote FastGS-Base and SprintGS-Base.}
  \label{fig:supp_breakdown}
\end{figure}

SprintGS-Base trains $2.15\times$ faster than FastGS-Base~\cite{FastGS} (Tab.~2 of the main paper).
\Cref{fig:supp_breakdown} shows where this speed-up comes from
by adding SprintGS-Base to the training-time breakdown of Fig.~1 of the main paper.
Since the GPU runs asynchronously to the CPU, a timer around a step measures only the launch of its kernels;
as in Fig.~1, we therefore synchronize the GPU after each step to measure its time.
This removes the overlap between the CPU and the GPU and slightly lengthens the training,
so Tab.~2 of the main paper, which reports the actual training time, synchronizes only at both ends of the training loop,
and we report the shares of the total time rather than the absolute times.
Since the total training time is roughly halved, a step whose share does not grow is shortened at least as much as the whole training.
This holds for the loss, whose share decreases (23\% $\to$ 17\%),
and for the backward rasterization, whose share stays almost unchanged (41\% $\to$ 40\%).
As a result, the forward rasterization now takes a noticeably larger share (23\% $\to$ 31\%),
which suggests that reducing its cost is an important direction for further acceleration.

\subsection{Cumulative Ablation}
\label{sec:supp_cum}

%%% Table: cumulative ablation
\begin{table}[t]
  \centering
  \footnotesize
  \setlength{\tabcolsep}{3pt}
  \caption{Cumulative ablation of SprintGS-Base.
  Top: the components are added to FastGS-Base~\cite{FastGS} one at a time.
  Bottom: the implementation optimizations adapted from Faster-GS~\cite{FasterGS} (F-E) are applied to FastGS-Base first, and F-A to F-D are then added.
  Time is the training time in seconds.}
  \label{tab:supp_cum}
  \begin{tabular}{lcccr}
    \toprule
    Method & PSNR$\uparrow$ & SSIM$\uparrow$ & LPIPS$\downarrow$ & Time$\downarrow$ \\
    \midrule
    FastGS-Base~\cite{FastGS}           & 27.31 & 0.818 & 0.256 & 172.0 \\
    \quad + F-A                         & 27.29 & 0.818 & 0.257 & 159.0 \\
    \quad + F-B                         & 27.38 & 0.822 & 0.252 & 152.8 \\
    \quad + F-C                         & 27.37 & 0.822 & 0.252 & 134.5 \\
    \quad + F-D                         & 27.38 & 0.822 & 0.252 & 110.7 \\
    \quad + F-E (= SprintGS-Base)       & 27.38 & 0.822 & 0.252 &  80.0 \\
    \midrule
    FastGS-Base~\cite{FastGS}           & 27.31 & 0.818 & 0.256 & 172.0 \\
    \quad + F-E                         & 27.31 & 0.818 & 0.256 & 142.5 \\
    \quad + F-A to F-D (= SprintGS-Base) & 27.38 & 0.822 & 0.252 &  80.0 \\
    \bottomrule
  \end{tabular}
\end{table}

Table~4 of the main paper removes one component at a time from the full model.
It shows how much each component saves when all the others are in place,
but not how the speed-up over FastGS-Base~\cite{FastGS} builds up,
nor how much of it comes from the implementation optimizations adapted from Faster-GS~\cite{FasterGS} (F-E).
We therefore add the components to FastGS-Base one at a time (\cref{tab:supp_cum}),
keeping all other settings as in the main paper.

\paragraph{Adding the components in order.}
The top block in \cref{tab:supp_cum} adds F-A to F-E.
Each added component shortens the training.
The time saved by F-A and F-B here (13.0\,s and 6.2\,s) is larger than in Tab.~4 of the main paper (9.3\,s and 1.8\,s), because they are added before F-C to F-E, when the rasterization and the per-Gaussian operations that they reduce are still slower.

\paragraph{Speed-up beyond the Faster-GS optimizations.}
The bottom block applies F-E to FastGS-Base first.
F-E alone shortens the training from 172.0\,s to 142.5\,s ($1.21\times$).
Adding F-A to F-D on top of it shortens the training to 80.0\,s,
a further $1.78\times$ speed-up out of the overall $2.15\times$.
The speed-up of SprintGS is therefore not mainly due to the optimizations adapted from Faster-GS, but to the new components F-A to F-D.
%% Section A.7
\subsection{Rendering Quality on DL3DV-10K}
\label{sec:supp_dl3dv}

\begin{figure*}[t]
  \centering
  % Extension omitted: uses .pdf with pdflatex (convert the .eps beforehand), .eps with latex
  \includegraphics[width=0.90\linewidth]{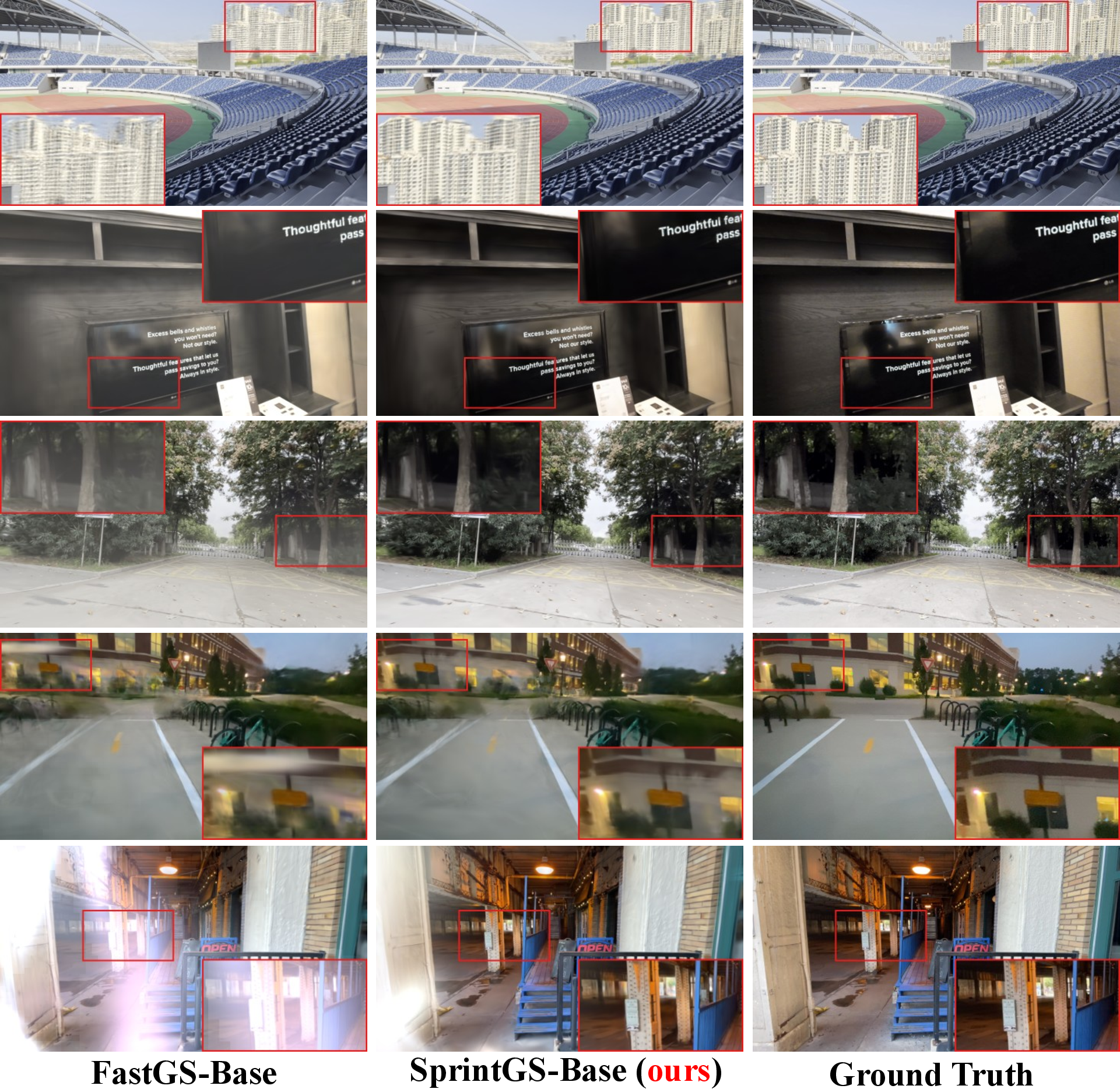}
  \caption{Rendering results on the 1K subset of DL3DV-10K~\cite{DL3DV-10K}: five scenes (top to bottom), one test view each.
  Columns: FastGS-Base, SprintGS-Base (ours), and the ground truth.
  The inset magnifies the region in the red box.}
  \label{fig:supp_dl3dv}
\end{figure*}

Due to space limitations, the rendering results on DL3DV-10K~\cite{DL3DV-10K}, evaluated in Sec.~5.2, are shown here.
\Cref{fig:supp_dl3dv} shows test views of five scenes, rendered by models trained under the same setting as Tab.~3.
SprintGS-Base trains 2.24$\times$ faster (141.8\,s $\to$ 63.3\,s)
and often reproduces fine structures more faithfully than FastGS-Base,
which is consistent with its slightly higher scores in Tab.~3.

% ===================================================
\section{Details of the Combinations with Other Fast Methods}
\label{sec:supp_other_baselines}

This section gives the implementation details of the experiment in Sec.~5.4 and Tab.~5.
For each method, we keep its own training strategy and add all components of SprintGS
with the same hyperparameters as SprintGS-Base (Secs.~4.1--4.5); the subsections below describe only the method-specific integration.

\subsection{Taming-3DGS + SprintGS}
\label{sec:supp_taming}

\noindent\textbf{Baseline.} Taming-3DGS~\cite{Taming3DGS} limits the number of Gaussians to a per-scene budget:
at each densification step, it clones or splits the Gaussians in descending order of its importance score
until their number reaches a target that grows toward the budget.
We use the budget mode of the official implementation with the per-scene budgets provided by the authors,
which is also the configuration of Tabs.~2 and 5 of the main paper.
The combination keeps this densification, its schedule, and its hyperparameters
(\eg, densification every 500 iterations until iteration 15,000).

\noindent\textbf{Integration of the components.}
All components are used without modification except for F-B,
and the importance score is accumulated in the forward pass of the SprintGS rasterizer.
F-B prunes after the densification of Taming-3DGS in each iteration where both run,
and at the next densification step, $\rho$ times the number of pruned Gaussians are cloned or split
in addition to the target of Taming-3DGS.
Since Taming-3DGS lacks the error-based candidates of FastGS from which SprintGS-Base takes these Gaussians,
we take them from the Gaussians that Taming-3DGS does not densify
(\ie, those whose positional gradient is below its threshold), in descending order of the importance score.

\subsection{DashGaussian + SprintGS}
\label{sec:supp_dash}

\noindent\textbf{Baseline.} DashGaussian~\cite{DashGaussian} accelerates 3DGS training with two schedulers of its own:
a resolution scheduler, which renders at a reduced resolution early in training and raises it according to the frequency content of the training images,
and a primitive scheduler, which controls the growth of the number of Gaussians.
It also prunes the Gaussians with an opacity below 0.005 at each densification step.
As the baseline, we use the sparse-Adam configuration, the fastest one of the official implementation (\textit{3DGS-fast-Dash} in its repository), which is also the configuration of Tabs.~2 and 5 of the main paper.
Since DashGaussian builds on 3DGS~\cite{Vanilla3DGS} rather than on FastGS~\cite{FastGS}, the combination keeps the densification and pruning of 3DGS and DashGaussian
with the hyperparameters of DashGaussian (\eg, its densification interval and gradient threshold),
and enables all components of SprintGS with the same hyperparameters as SprintGS-Base.

\noindent\textbf{Integration of the components.}
Scheduled tile culling (F-A) replaces the tile assignment of DashGaussian without modification:
since $\tau$ is a threshold on alpha, it does not depend on the image resolution.
Convergence-aware pruning (F-B) runs until the end of the densification of DashGaussian at iteration 27,000 instead of 15,000.
In each iteration where both run, the pruning and densification of DashGaussian run first, and F-B prunes afterwards.
For the shuffle-efficient backward pass (F-C), the official rasterizer of DashGaussian passes 11 values between threads,
including the inverse depth and its gradient, although depth supervision is not used in this configuration
(Tab.~1 of the main paper); the combination passes only 2.
The single-launch Adam (F-E2) replaces the sparse Adam of DashGaussian,
so the combination updates all Gaussians at every step, as the standard Adam of 3DGS does.

\subsection{SkipGS + SprintGS}
\label{sec:supp_skipgs}

\noindent\textbf{Baseline.} SkipGS~\cite{SkipGS} is a plug-in for 3DGS training whose authors provide an official integration with FastGS~\cite{FastGS},
which we use for the SkipGS results in the main paper and as the baseline of this combination.

\noindent\textbf{Integration of the components.} We replace FastGS in this integration with SprintGS-Base, keeping the SkipGS controller and its settings unchanged
at the same place in the training loop (between the loss and the backward pass)
and SprintGS-Base with the same hyperparameters as in the main paper.
The two methods are complementary:
SkipGS reduces the \emph{number} of backward passes by skipping, after densification, those of views whose loss is at or below its running average,
whereas SprintGS reduces the cost of \emph{each} training step.

\subsection{Differences in the Speed-ups}
\label{sec:supp_speedup_diff}
SprintGS accelerates FastGS-Base, Taming-3DGS, DashGaussian, and SkipGS by
$2.15\times$, $2.76\times$, $1.73\times$, and $2.12\times$, respectively (Tabs.~2 and 5).
Here, we discuss why the speed-ups for the other fast methods differ from that for FastGS-Base.

\noindent\textbf{Taming-3DGS.}
Two differences from FastGS lead to a speed-up above that over FastGS-Base.
First, the rasterizer of Taming-3DGS assigns tiles by the $3\sigma$ bounding square of 3DGS~\cite{Vanilla3DGS};
replacing it with F-A also brings the ellipse-based tile assignment of Speedy-Splat~\cite{SpeedySplat}, which FastGS already uses and which assigns each Gaussian to fewer tiles than the bounding square.
Second, Taming-3DGS updates all parameters, including the SH coefficients, at every iteration,
whereas the official implementation of FastGS~\cite{FastGS} updates some of them less often (\eg, the SH coefficients every 32 to 64 iterations after iteration 15,000).
The optimizer thus takes a larger share of the training time, and F-E2 removes more of it.

\noindent\textbf{DashGaussian.}
DashGaussian+SprintGS keeps 1.76\,M Gaussians, about 5.5 times as many as SprintGS-Base (0.32\,M; Tabs.~2 and 5).
The operations whose cost grows with the number of Gaussians and that SprintGS does not target thus take a larger share of the training time,
and SprintGS, designed for compact models such as FastGS, does not reach a $2\times$ speed-up.

\noindent\textbf{SkipGS.}
SkipGS builds on FastGS and keeps a model of nearly the same size,
so the combination keeps nearly the same speed-up as SprintGS-Base over FastGS-Base.

% ===================================================
\section{Details of the Extensions to Other Reconstruction Tasks}
\label{sec:supp_extension}
This section gives the datasets, baselines, and integration details of the experiments in Sec.~5.5
and Tabs.~6--8, together with additional rendering results.

\begin{figure*}[t]
  \centering
  \includegraphics[width=0.92\linewidth]{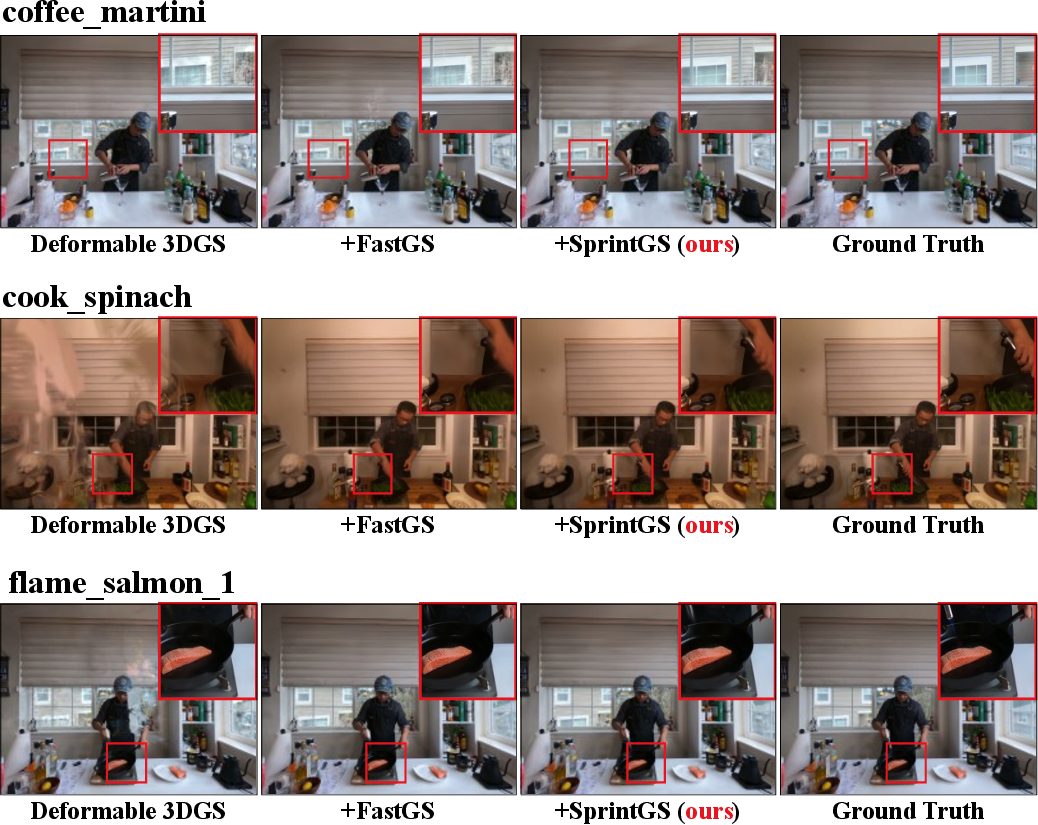}
  \caption{Dynamic scene reconstruction on Neu3D~\cite{Neu3D}: \textit{coffee\_martini}, \textit{cook\_spinach}, and \textit{flame\_salmon\_1} (top to bottom), test camera cam00.
  The inset magnifies the region in the red box.}
  \label{fig:supp_dynamic}
\end{figure*}

\subsection{Dynamic Scene Reconstruction}
\label{sec:supp_dynamic}

\noindent\textbf{Dataset and protocol.}
We use the six scenes of the Neu3D dataset~\cite{Neu3D}.
Following the default of the \textit{fast-d3dgs} branch of the official FastGS code~\cite{FastGS},
we take the first 24 frames of every camera,
downscale them from $2704\times2028$ to $1600\times1200$, and hold out cam00 for testing,
which gives 408 to 480 training images (17 to 20 cameras, depending on the scene) and 24 test images per scene.

\noindent\textbf{Baselines.}
Deformable-3DGS~\cite{Deformable3DGS} is run with its official code and default settings.
For Deformable-3DGS+FastGS, we use the \textit{fast-d3dgs} branch,
which combines the deformation network of Deformable-3DGS with FastGS.

\noindent\textbf{Integration into SprintGS.}
We add the components of SprintGS to the \textit{fast-d3dgs} branch~\cite{FastGS}
and keep the hyperparameters of Deformable-3DGS+FastGS, so the two differ only in the components of SprintGS.
All components are enabled except F-E1:
the deformation network adds its offsets to the position, rotation, and scale after activation in PyTorch,
which F-E1 bypasses by applying the activations inside the rasterizer.
F-A and F-B keep the settings of the static experiments.

\noindent\textbf{Rendering results.}
\Cref{fig:supp_dynamic} shows the rendering results on the test camera.

\subsection{Sparse-View Reconstruction}
\label{sec:supp_sparse}

\begin{figure*}[t]
  \centering
  \includegraphics[width=0.92\linewidth]{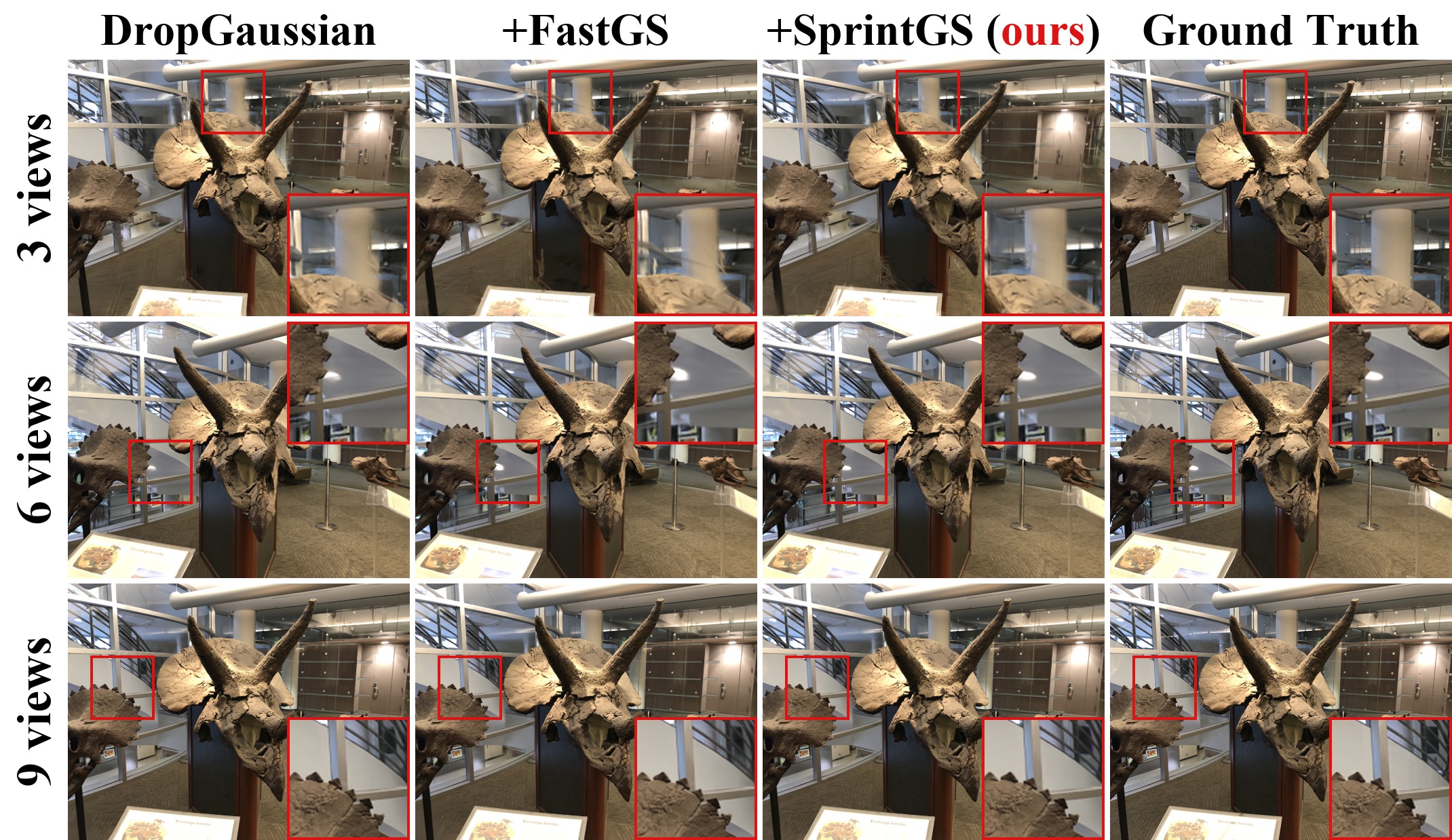}
  \caption{Sparse-view reconstruction on \textit{horns} (LLFF~\cite{LLFF}) with 3, 6, and 9 training views (top to bottom).
  The inset magnifies the region in the red box.}
  \label{fig:supp_sparse_horns}
\end{figure*}

\begin{figure*}[t]
  \centering
  \includegraphics[width=0.92\linewidth]{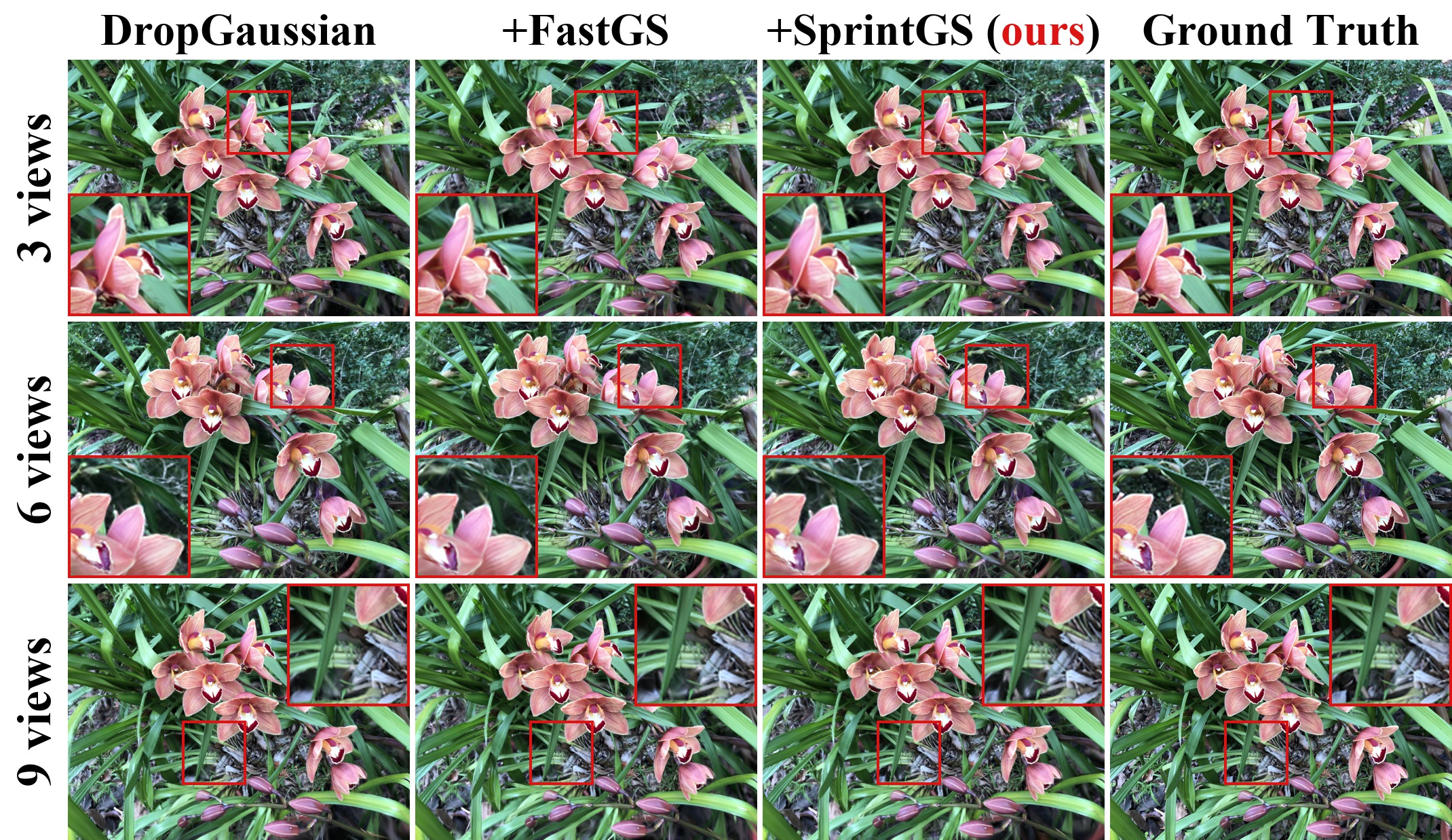}
  \caption{Sparse-view reconstruction on \textit{orchids} (LLFF~\cite{LLFF}) with 3, 6, and 9 training views (top to bottom).
  The inset magnifies the region in the red box.}
  \label{fig:supp_sparse_orchids}
\end{figure*}

\noindent\textbf{Dataset and protocol.}
Following DropGaussian~\cite{DropGaussian}, we use the eight scenes of LLFF~\cite{LLFF} downscaled by 8 ($504\times378$),
hold out every eighth image for testing, and sample 3, 6, or 9 training views uniformly from the rest.

\noindent\textbf{Baselines.}
DropGaussian is run with its official code and default settings,
and DropGaussian+FastGS with the LLFF script of the sparse-view branch of the official FastGS code~\cite{FastGS}.
Both train for 10,000 iterations and densify until the end of training,
every 100 (DropGaussian) or 500 iterations (DropGaussian+FastGS).

\noindent\textbf{Integration into SprintGS.}
We add the components of SprintGS to the sparse-view branch~\cite{FastGS}
and keep the hyperparameters of DropGaussian+FastGS, so the two differ only in the components of SprintGS.
All components are enabled except F-E1:
DropGaussian modifies the activated opacity in PyTorch, which F-E1 bypasses by applying the activations inside the rasterizer.
F-A follows Eq.~(6) of the main paper with $T=10{,}000$ and thus $\Delta=1{,}000$,
and F-B keeps the settings of the static experiments and runs until the end of densification.

\noindent\textbf{Rendering results.}
\Cref{fig:supp_sparse_horns,fig:supp_sparse_orchids} show test views of \textit{horns} and \textit{orchids}.
SprintGS renders them with quality comparable to or better than the baselines,
consistent with Tab.~7 of the main paper, where it gives the best PSNR with the fewest Gaussians for every number of views.

\subsection{Surface Reconstruction}
\label{sec:supp_surface}

\begin{figure*}[t]
  \centering
  \includegraphics[width=0.75\linewidth]{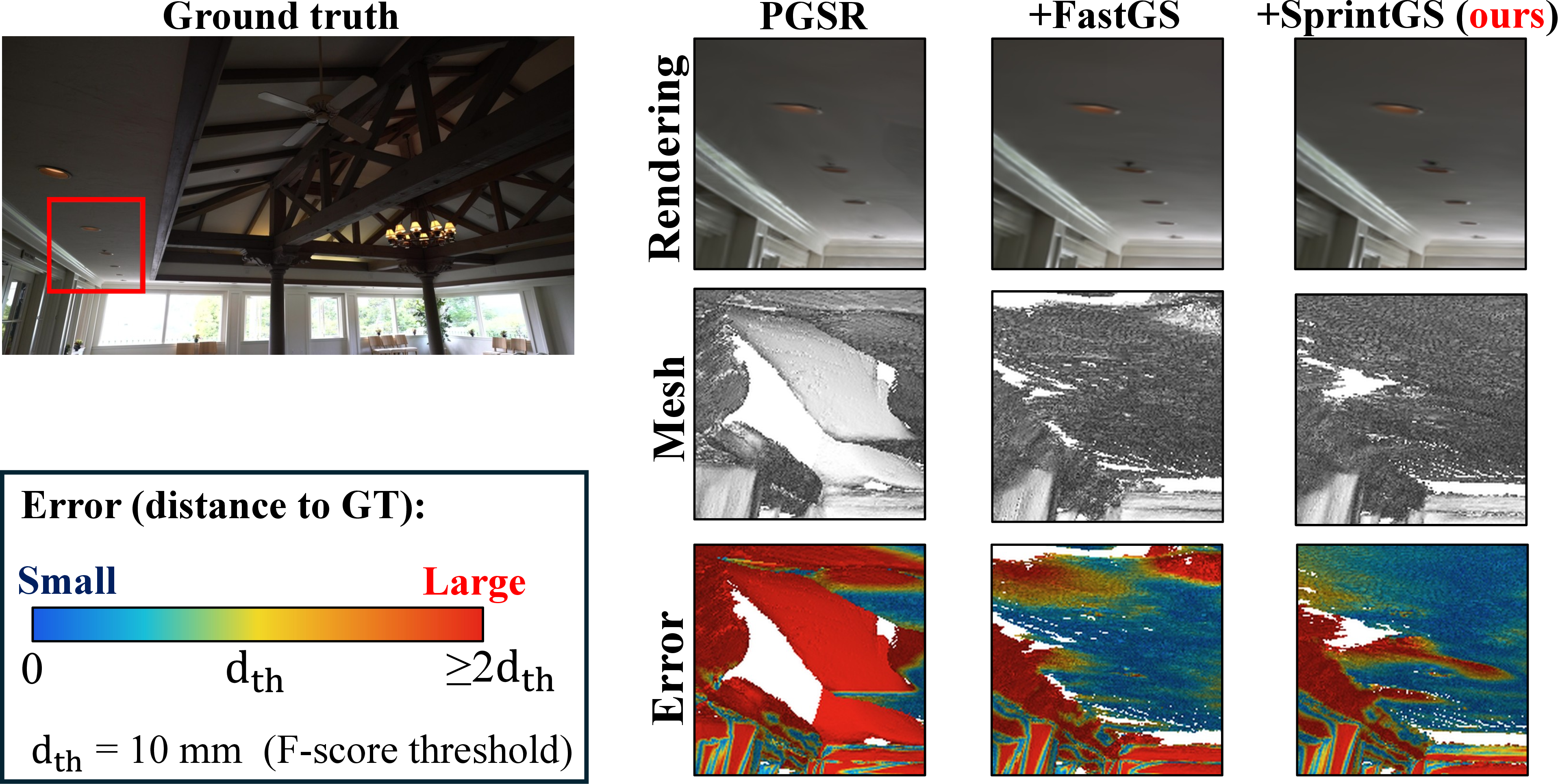}
  \caption{Surface reconstruction on \textit{meetingroom} of Tanks \& Temples~\cite{TT}. The ground-truth image (left) shows the magnified region in the red box. For PGSR~\cite{PGSR}, PGSR+FastGS, and PGSR+SprintGS (ours), the rows show the rendering, the extracted mesh, and its distance to the ground-truth laser scan (Error), where $d_{\mathrm{th}}$ (10\,mm) is the distance threshold of the F-score of Tanks \& Temples~\cite{TT}.}
  \label{fig:supp_surface}
\end{figure*}

\noindent\textbf{Dataset and protocol.}
We use four scenes of the Tanks \& Temples dataset~\cite{TT}: \textit{truck}, \textit{caterpillar}, \textit{ignatius}, and \textit{meetingroom}.
The training and evaluation follow PGSR~\cite{PGSR} and PGSR+FastGS:
after training, a mesh is extracted by truncated signed distance function (TSDF) fusion and evaluated by the F-score of the official Tanks \& Temples evaluation,
and PSNR, SSIM, and LPIPS are measured on the test images.

\noindent\textbf{Baselines.}
PGSR~\cite{PGSR} is run with its official code and its default settings for Tanks \& Temples.
For PGSR+FastGS, we use the code released with FastGS~\cite{FastGS} under the name \textit{Fast-PGSR},
which combines the planar rendering and the losses of PGSR with FastGS.

\noindent\textbf{Integration into SprintGS.}
We add the components of SprintGS to the PGSR+FastGS code
and keep its hyperparameters, so the two differ only in the components of SprintGS.
All components (F-A to F-E) are enabled.
PGSR+FastGS uses two rasterizers:
the planar rasterizer of PGSR, which renders the color and the planar maps (normal and plane distance) in one pass for the training loss,
and the color rasterizer of FastGS, which renders the sampled views for the multi-view scores that drive the densification and pruning of FastGS.
F-A determines the tiles covered by each Gaussian in both rasterizers.

\noindent\textbf{Rendering results.}
\Cref{fig:supp_surface} shows the reconstructed surface on \textit{meetingroom}:
like PGSR+FastGS, PGSR+SprintGS suppresses the floating surface that PGSR produces below the ceiling,
and most of its surface lies within $d_{\mathrm{th}}$ of the ground truth.

% ===================================================
\section{Details and Analysis of the Components}
\label{sec:supp_components}
This section gives, for each component of SprintGS (Secs.~4.1--4.5),
the details that are omitted from the main text due to space limitations,
followed by analyses of the component, such as its sensitivity to the hyperparameters.

\subsection{Scheduled Tile Culling (F-A)}
\label{sec:supp_fa}

Scheduled tile culling (F-A, Sec.~4.1 of the main paper) keeps the culling threshold $\tau$ at a coarse value $\tau_c$ for most of training
and then decreases it to $\epsilon\,(=1/255)$, which gives the exact footprint.
We first detail the logistic function of its transition and then analyze its sensitivity to $\tau_c$.

%% ------------------------------------------------------------
\subsubsection{Logistic Transition}
\label{sec:supp_fa_logistic}

The transition of $\tau(t)$ in Eq.~(6) of the main paper uses a normalized logistic function
\begin{equation}
  s(u)=\frac{l(u)-l(0)}{l(1)-l(0)},\qquad
  l(x)=\frac{1}{1+\exp\bigl(-k\,(x-0.5)\bigr)},
\end{equation}
where $u\in[0,1)$ is the progress within the transition and $k$ controls its steepness.
The normalization gives $s(0)=0$ and $s(1)=1$, so $\tau$ starts at $\tau_c$ and reaches $\epsilon$ at the end of the transition.
Since $\tau=\tau_c\,(\epsilon/\tau_c)^{s(u)}$, $\tau$ follows $s$ in log scale:
it decreases slowly at the beginning and the end of the transition and fastest in the middle,
and the decrease becomes geometric as $k\to0$.
We use $k=10$, $\tau_c=0.10$, and $\Delta=T/10=3{,}000$ iterations for all scenes and both SprintGS-Base and SprintGS-Big,
so that $\tau=\tau_c$ for $t<24{,}000$, $\tau$ decreases to $\epsilon$ for $24{,}000\le t<27{,}000$, and $\tau=\epsilon$ for $t\ge27{,}000$.

%% ------------------------------------------------------------
\subsubsection{Sensitivity to $\tau_c$}
\label{sec:supp_sensitivity}

We vary $\tau_c$ in SprintGS-Base and keep all other components and hyperparameters unchanged. For reference, we also report three tile assignments without the schedule: the exact footprint of Speedy-Splat~\cite{SpeedySplat} throughout training ($\tau_c=\epsilon$); the uniformly shrunk footprint (compact box) of FastGS~\cite{FastGS}, which SprintGS-Base uses without F-A (Tab.~4 of the main paper) and which shrinks the exact footprint by a factor $\beta$ set per dataset ($0.5$ on Mip-NeRF 360~\cite{MipNeRF360} and $0.7$ on Tanks \& Temples~\cite{TT} and Deep Blending~\cite{DeepBlending});
and $\tau=0.10$ kept until the end of training and also used for rendering. \Cref{tab:supp_tau_c} reports the averages over all 13 scenes.

\begin{table}[t]
  \centering
  \footnotesize
  \setlength{\tabcolsep}{2pt}
  \caption{Sensitivity to $\tau_c$ of F-A in SprintGS-Base (averages over all 13 scenes). The top block gives the tile assignments without the schedule of F-A.}
  \label{tab:supp_tau_c}
  \begin{tabular}{lccccc}
    \toprule
    Tile assignment & PSNR$\uparrow$ & SSIM$\uparrow$ & LPIPS$\downarrow$ & Time$\downarrow$ & \#G$\downarrow$\\
    \midrule
    Exact footprint~\cite{SpeedySplat} & 27.45 & 0.822 & 0.252 & 112.2 & 0.313\\
    Compact box (FastGS)~\cite{FastGS} & 27.39 & 0.822 & 0.251 &  89.3 & 0.317\\
    Fixed $\tau=0.10$ (no schedule)    & 27.25 & 0.820 & 0.252 &  77.6 & 0.318\\
    \midrule
    F-A, $\tau_c=0.01$    & 27.44 & 0.823 & 0.251 & 104.3 & 0.319\\
    F-A, $\tau_c=0.025$   & 27.44 & 0.823 & 0.251 &  95.7 & 0.323\\
    F-A, $\tau_c=0.05$    & 27.41 & 0.823 & 0.251 &  88.0 & 0.322\\
    F-A, $\tau_c=0.075$   & 27.40 & 0.822 & 0.252 &  83.8 & 0.320\\
    F-A, $\tau_c=0.10$ (default) & 27.38 & 0.822 & 0.252 &  80.0 & 0.318\\
    F-A, $\tau_c=0.125$   & 27.33 & 0.821 & 0.253 &  79.0 & 0.315\\
    F-A, $\tau_c=0.15$    & 27.31 & 0.821 & 0.254 &  77.5 & 0.313\\
    F-A, $\tau_c=0.20$    & 27.26 & 0.820 & 0.256 &  75.4 & 0.307\\
    \bottomrule
  \end{tabular}
\end{table}

\noindent\textbf{Results.}
A smaller $\tau_c$ trades speed for quality:
from $\tau_c=0.20$ to $0.01$, PSNR rises from 27.26 to 27.44\,dB and the training time from 75.4 to 104.3\,s.
The schedule is what makes a coarse $\tau$ affordable:
with the same $\tau_c=0.10$, restoring the exact footprint in the last 10\% of the iterations
raises PSNR by 0.13\,dB (27.25 $\to$ 27.38\,dB) at a cost of only 2.4\,s (77.6 $\to$ 80.0\,s).
Compared with the shrunk footprint of FastGS, F-A is faster at nearly the same quality for $\tau_c$ between 0.05 and 0.10,
and both faster and higher PSNR at $\tau_c=0.05$ and $0.075$.
Moreover, F-A uses a single value, $\tau_c=0.10$, on all datasets,
whereas the shrunk footprint requires a per-dataset $\beta$.

\subsection{Convergence-Aware Pruning (F-B)}
\label{sec:supp_fb}

Convergence-aware pruning (F-B, Sec.~4.2 of the main paper)
prunes the Gaussians that satisfy both conditions of Eq.~(7) of the main paper:
a low opacity ($o_i<\theta_o$) and a small convergence signal ($v_i\le Q_q(v)$),
where $v_i$ is the Adam second moment of the position summed over its three components.
The test runs every $K$ iterations, and the freed budget is partly re-allocated to poorly reconstructed regions with the ratio $\rho$. 
We analyze its design choices and its sensitivity to the hyperparameters in SprintGS-Base.

%% ------------------------------------------------------------
\subsubsection{Design Choices}
\label{sec:supp_fb_design}

We change one part of the convergence test
and keep all other components and hyperparameters unchanged
($\theta_o=0.1$, $q=0.5$, $K=500$, $\rho=0.4$).
\begin{itemize}[leftmargin=1.2em,itemsep=1pt,topsep=2pt,parsep=0pt]
  \item \textbf{Condition.} We use only one of the two conditions:
  the opacity condition alone ($q=1$) or the convergence condition alone ($\theta_o=1$).
  \item \textbf{Moment.} We replace the second moment with the first moment of the position,
  \ie, $v_i$ is the sum of the absolute values of the three components of the running mean of the gradient
  instead of that of the squared gradient.
  \item \textbf{Parameter.} We compute $v_i$ from the second moment of another parameter:
  the scale, the rotation, or the base color (the zeroth-order spherical harmonics coefficients).
\end{itemize}
\Cref{tab:supp_fb_design} reports the averages over all 13 scenes.

%%% Table: design choices of F-B
\begin{table}[t]
  \centering
  \footnotesize
  \setlength{\tabcolsep}{2pt}
  \caption{Design choices of F-B in SprintGS-Base (averages over all 13 scenes).
  $^\ast$Training fails on 9 of the 13 scenes because almost all Gaussians are removed.}
  \label{tab:supp_fb_design}
  \begin{tabular}{llccccc}
    \toprule
    & Variant & PSNR$\uparrow$ & SSIM$\uparrow$ & LPIPS$\downarrow$ & Time$\downarrow$ & \#G$\downarrow$\\
    \midrule
    \multicolumn{2}{l}{Full (opacity $\wedge$ convergence)}
                                  & 27.38 & 0.822 & 0.252 & 80.0 & 0.318 \\
    \midrule
    Condition & Opacity only      & 27.30 & 0.821 & 0.254 & 79.9 & 0.313 \\
              & Convergence only  & \multicolumn{5}{c}{fails$^\ast$} \\
    \midrule
    Moment    & 1st moment        & 27.33 & 0.822 & 0.252 & 79.8 & 0.316 \\
    \midrule
    Parameter & Scale             & 27.38 & 0.822 & 0.253 & 80.2 & 0.314 \\
              & Rotation          & 27.38 & 0.822 & 0.253 & 80.3 & 0.317 \\
              & Base color        & 27.38 & 0.822 & 0.253 & 80.0 & 0.316 \\
    \bottomrule
  \end{tabular}
\end{table}

\noindent\textbf{Both conditions are needed.}
With the opacity condition alone, F-B removes more Gaussians than the full model,
and PSNR drops by 0.08\,dB.
The additional Gaussians are transparent but still being updated,
and some of them may still become opaque (Sec.~4.2 of the main paper).
With the convergence condition alone, F-B also removes opaque Gaussians that have converged to a useful state.
Since $Q_{0.5}(v)$ is the median, this condition selects half of the Gaussians at every test,
so the model shrinks rapidly, and training fails on 9 of the 13 scenes because almost no Gaussians are left; the remaining four scenes reach only 14.7\,dB.

\noindent\textbf{Second moment.}
With the first moment, PSNR drops by 0.05\,dB.
The first moment averages the gradient with its sign, so gradients that change direction
from view to view cancel out, and a Gaussian that is still being updated can have a small first moment.
The second moment averages the squared gradient without this cancellation
and thus better identifies Gaussians that the loss has stopped moving.

\noindent\textbf{Choice of the parameter.}
The second moments of the scale, the rotation, and the base color give the same PSNR as that of the position (27.38\,dB)
with a similar number of Gaussians.
The convergence test is thus not sensitive to which parameter it observes.

%% ------------------------------------------------------------
\subsubsection{Sensitivity to Hyperparameters}
\label{sec:supp_fb_sens}

We vary each hyperparameter of F-B while keeping the others at their defaults:
the quantile $q$ of the convergence test (default 0.5),
the re-allocation ratio $\rho$ (0.4),
the opacity threshold $\theta_o$ (0.1),
and the interval $K$ of the test (500 iterations).
\Cref{tab:supp_fb_sens} reports the averages over all 13 scenes.

%%% Table: sensitivity of F-B
\begin{table}[t]
  \centering
  \footnotesize
  \setlength{\tabcolsep}{3pt}
  \caption{Sensitivity to the hyperparameters of F-B in SprintGS-Base
  (averages over all 13 scenes).
  One hyperparameter is changed at a time, and the others are kept at their defaults.}
  \label{tab:supp_fb_sens}
  \begin{tabular}{llccccc}
    \toprule
    & Value & PSNR$\uparrow$ & SSIM$\uparrow$ & LPIPS$\downarrow$ & Time$\downarrow$ & \#G$\downarrow$\\
    \midrule
    $q$ & 0.25          & 27.36 & 0.822 & 0.251 & 82.7 & 0.329\\
        & 0.5 (default) & 27.38 & 0.822 & 0.252 & 80.0 & 0.318\\
        & 0.75          & 27.36 & 0.821 & 0.253 & 80.4 & 0.313\\
    \midrule
    $\rho$ & 0.0           & 27.25 & 0.818 & 0.260 & 77.6 & 0.284\\
           & 0.1           & 27.28 & 0.819 & 0.258 & 77.8 & 0.291\\
           & 0.2           & 27.31 & 0.820 & 0.256 & 79.2 & 0.299\\
           & 0.3           & 27.34 & 0.821 & 0.254 & 79.5 & 0.308\\
           & 0.4 (default) & 27.38 & 0.822 & 0.252 & 80.0 & 0.318\\
           & 0.5           & 27.40 & 0.823 & 0.250 & 81.9 & 0.329\\
    \midrule
    $\theta_o$ & 0.05          & 27.38 & 0.822 & 0.251 & 83.6 & 0.341\\
               & 0.1 (default) & 27.38 & 0.822 & 0.252 & 80.0 & 0.318\\
               & 0.15          & 27.32 & 0.821 & 0.255 & 78.3 & 0.294\\
               & 0.2           & 27.24 & 0.819 & 0.259 & 76.3 & 0.270\\
    \midrule
    $K$ & 250           & 27.37 & 0.822 & 0.253 & 79.9 & 0.310\\
        & 500 (default) & 27.38 & 0.822 & 0.252 & 80.0 & 0.318\\
        & 1000          & 27.42 & 0.823 & 0.250 & 82.7 & 0.337\\
    \bottomrule
  \end{tabular}
\end{table}

\noindent\textbf{Results.}
F-B is insensitive to $q$: from 0.25 to 0.75, PSNR changes by at most 0.02\,dB.
The re-allocation ratio $\rho$ and the opacity threshold $\theta_o$ control the trade-off between quality and speed through the number of Gaussians.
A larger $\rho$ re-allocates more of the freed budget:
from $\rho=0$ to $0.5$, the number of Gaussians increases from 0.284\,M to 0.329\,M,
and PSNR, SSIM, and LPIPS improve monotonically (27.25 to 27.40\,dB), while training becomes slower (77.6\,s to 81.9\,s).
Pruning without re-allocation ($\rho=0$) is 0.04\,dB below SprintGS-Base without F-B
(27.29\,dB, Tab.~4 of the main paper),
which shows that the re-allocation is necessary for the quality gain of F-B.
A larger $\theta_o$ prunes more Gaussians and makes training faster at a lower PSNR,
and a larger $K$ prunes less often and has the opposite effect.

\subsection{Shuffle-Efficient Backward Pass (F-C)}
\label{sec:supp_fc}

%%% Table: ablation of F-C
\begin{table}[t]
  \centering
  \footnotesize
  \setlength{\tabcolsep}{1.5pt}
  \caption{Effect of each step of F-C in SprintGS-Base (averages over all 13 scenes).
  \#Sh is the number of values shuffled at every step,
  and Sh.\ mem.\ indicates that shared memory is used: for the per-pixel constants (F-C variants) or for the bucket start state (control).
  The bottom row is a control that uses shared memory without reducing the shuffled values.}
  \label{tab:supp_fc_abl}
  \begin{tabular}{lcccccc}
    \toprule
    Variant & Sh.\ mem. & \#Sh & PSNR$\uparrow$ & SSIM$\uparrow$ & LPIPS$\downarrow$ & Time$\downarrow$ \\
    \midrule
    Shuffle all            &            & 9 & 27.37 & 0.822 & 0.252 & 98.2 \\
    \,+ Pixel window       & \checkmark & 4 & 27.38 & 0.822 & 0.252 & 86.0 \\
    \,+ Scalar state (ours, F-C) & \checkmark & 2 & 27.38 & 0.822 & 0.252 & 80.0 \\
    \midrule
    Shuffle all + staged state & \checkmark & 9 & 27.37 & 0.822 & 0.252 & 93.2 \\
    \bottomrule
  \end{tabular}
\end{table}

The shuffle-efficient backward pass (F-C, Sec.~4.3 of the main paper)
reduces the values passed between threads by warp shuffle at every step of the splat-parallel backward pass
from 9 to 2 (Tab.~1 of the main paper).
We first give the pseudocode for one bucket and then separate the effects of its two steps.

%% ------------------------------------------------------------
\subsubsection{Backward Pass of One Bucket}
\label{sec:supp_fc_bucket}

\begin{algorithm}[t]
\caption{Backward pass of one bucket in F-C (one warp).
Thread $k$ holds the $k$-th Gaussian of the bucket, and the 256 pixels of the tile
pass through the threads one step at a time.
Only $T$ and $s$ are shuffled.}
\label{alg:fc_backward}
\small
\begin{algorithmic}[1]
\Require per pixel $x$: $\mathbf{g}_x$, $T_{\mathrm{fin},x}$, $i_{\mathrm{last},x}$, and color $\mathbf{C}_x$ without the background;
transmittance $\hat{T}_x$ and color $\hat{\mathbf{A}}_x$ accumulated in front of the bucket (from the forward pass);
background color $\mathbf{b}$
\For{$m=0,\dots,286$}
  \If{$m \bmod 32 = 0$} \Comment{for the next 32 pixels}
    \State load $\mathbf{g}_x, i_{\mathrm{last},x}$ to shared memory
    \State $s^0_x\gets(\mathbf{C}_x-\hat{\mathbf{A}}_x+T_{\mathrm{fin},x}\mathbf{b})\cdot\mathbf{g}_x$ to shared memory \Comment{background folded in}
  \EndIf
  \State $(T,s)\gets$ \textsc{ShuffleUp}$(T,s)$ \Comment{from thread $k{-}1$}
  \State $x\gets m-k$ \Comment{pixel held by thread $k$}
  \If{$k=0$} \Comment{start of the bucket}
    \State $T\gets\hat{T}_x$,\ \ $s\gets s^0_x$
  \EndIf
  \If{$0\le x<256$ \textbf{and} Gaussian $k$ is not behind $i_{\mathrm{last},x}$ \textbf{and} $\alpha_k\ge\epsilon$}
    \State $w\gets\mathbf{c}_k\cdot\mathbf{g}_x$ \Comment{$\mathbf{g}_x$ from shared memory}
    \State $s\gets s-\alpha_k T\,w$ \Comment{$=\mathbf{r}_k\cdot\mathbf{g}_x=s_k$}
    \State $\partial L/\partial\alpha_k \mathrel{+}= T\,w - s/(1-\alpha_k)$ \Comment{Eq.~(9)}
    \State $T\gets T\,(1-\alpha_k)$
  \EndIf
\EndFor
\State add the accumulated gradients to global memory
\end{algorithmic}
\end{algorithm}

\Cref{alg:fc_backward} gives the pseudocode of F-C for one bucket.
The bucket structure follows the splat-parallel backward pass of Taming-3DGS~\cite{Taming3DGS}, which FastGS~\cite{FastGS} also uses,
and the tile size follows 3DGS~\cite{Vanilla3DGS}; neither is part of our design.
The Gaussians of a tile, sorted front to back, are divided into \emph{buckets} of 32 consecutive Gaussians,
and each bucket is processed by one \emph{warp}, a group of 32 threads executed together,
with one Gaussian per thread.
At step $m$, thread $k$ holds pixel $x=m-k$,
so the 256 pixels of a $16\times16$ tile pass through the 32 threads in 287 steps.
The forward pass stores, for every bucket and pixel, the transmittance $\hat{T}_x$ and the
color $\hat{\mathbf{A}}_x$ accumulated in front of the bucket, from which thread 0 starts each pixel.
Since the stored color $\mathbf{C}_x$ excludes the background, which is seen through the final transmittance $T_{\mathrm{fin},x}$,
$(\mathbf{C}_x-\hat{\mathbf{A}}_x+T_{\mathrm{fin},x}\mathbf{b})\cdot\mathbf{g}_x$ equals $s$ in front of the first Gaussian of the bucket.
Folding the background into this start value, which is kept in the shared-memory window,
removes the background term $-T_{\mathrm{fin},x}\,(\mathbf{b}\cdot\mathbf{g}_x)/(1-\alpha_k)$ that 3DGS evaluates at every step,
so $T_{\mathrm{fin},x}$ is used only once per pixel when the window is loaded.
In F-C, the per-pixel constants are loaded every 32 steps into a ring window of 64 pixels in shared memory,
which covers the pixels held by the 32 threads at any step.
Each thread then updates $s$ and $T$ front to back by Eq.~(9) of the main paper and accumulates the gradient of its Gaussian; the other gradients (color, mean, and conic) are computed as in FastGS~\cite{FastGS}.

%% ------------------------------------------------------------
\subsubsection{Effect of Each Step}
\label{sec:supp_fc_abl}

F-C reduces the shuffled values in two steps:
\begin{itemize}[leftmargin=1.2em,itemsep=1pt,topsep=2pt,parsep=0pt]
  \item \textbf{Pixel window (9 $\to$ 4 values).} The per-pixel constants $\mathbf{g}_x$, $T_{\mathrm{fin},x}$, and $i_{\mathrm{last},x}$ (5 values)
  are taken from the shared-memory window instead of being shuffled,
  so only $T$ and the color $\mathbf{r}$ behind the Gaussian (4 values) are shuffled.
  \item \textbf{Scalar state (4 $\to$ 2 values).} The color $\mathbf{r}$ (3 values) is replaced by the scalar $s=\mathbf{r}\cdot\mathbf{g}_x$,
  so only $T$ and $s$ (2 values) are shuffled.
\end{itemize}
To separate the effect of the shared memory itself from that of the number of shuffled values,
we also evaluate a control that keeps the 9 shuffled values
but stages the bucket start state ($\hat{T}_x$ and $\hat{\mathbf{A}}_x$) in shared memory
instead of reading it from global memory by thread 0.

\noindent\textbf{Results.}
\Cref{tab:supp_fc_abl} reports the averages over all 13 scenes, with the training time in seconds.
All variants are built on SprintGS-Base with all other components and hyperparameters unchanged.
Both reductions of the shuffled values shorten training (98.2 $\to$ 86.0 $\to$ 80.0\,s), whereas the control, which also uses shared memory, saves only 5.0\,s.
Most of the gain of F-C thus comes from the reduction of the shuffled values, which is consistent with Sec.~4.3 of the main paper.

\subsection{Fused Loss with Forward-Pass Gradient (F-D)}
\label{sec:supp_fd}

This section details F-D (Sec.~4.4 of the main paper).
We use the notation of Eq.~(3) of the main paper:
$C_p$ and $C^{\mathrm{gt}}_p$ are the rendered and ground-truth values,
and $p$ and $q$ index the $M$ color values of an image.
F-D does not change the loss of 3DGS~\cite{Vanilla3DGS} and FastGS~\cite{FastGS}: it uses $\lambda=0.2$ and computes SSIM per channel with an $11\times11$ Gaussian window $\mathcal{W}$ (standard deviation 1.5),
zero padding outside the image, and the stabilizing constants $0.01^2$ and $0.03^2$.

\noindent\textbf{Implementation in FastGS.}
FastGS computes the L1 term with PyTorch operations (a subtraction, an absolute value, and a mean)
and SSIM with the \texttt{fused-ssim} extension of Taming-3DGS~\cite{Taming3DGS}.
In the forward pass, \texttt{fused-ssim} zero-initializes four image-sized buffers
and writes the SSIM map and three maps of its derivatives, which another kernel averages; further scalar operations combine the two terms.
In the backward pass, automatic differentiation expands the constant $\partial L/\partial\ell_p=1/M$ into an image-sized map,
the SSIM backward kernel convolves it with the three derivative maps,
and further kernels compute the L1 gradient and add the two gradients.
In total, the loss of FastGS launches 13 kernels in the forward pass and 10 in the backward pass in every training step.

\noindent\textbf{Gradient in closed form.}
By Eq.~(10) of the main paper,
\begin{equation}
  g_q=\frac{\partial L}{\partial C_q}
  =\frac{1-\lambda}{M}\,\mathrm{sign}(C_q-C^{\mathrm{gt}}_q)-\frac{\lambda}{M}\sum_p\frac{\partial\,\mathrm{SSIM}_p}{\partial C_q},
  \label{eq:supp_fd_chain}
\end{equation}
where $\partial\,\mathrm{SSIM}_p/\partial C_q$ is nonzero only when $q$ lies in the window of $p$.
$\mathrm{SSIM}_p$ is a function of the local statistics
$E^{C}=\mathcal{W}*C$, $E^{\mathrm{gt}}=\mathcal{W}*C^{\mathrm{gt}}$, $E^{CC}=\mathcal{W}*C^2$,
$E^{\mathrm{gt}\mathrm{gt}}=\mathcal{W}*(C^{\mathrm{gt}})^2$, and $E^{C\mathrm{gt}}=\mathcal{W}*(C\,C^{\mathrm{gt}})$,
and the rendering enters only through $E^{C}$, $E^{CC}$, and $E^{C\mathrm{gt}}$.
Let $S^{C}_p$, $S^{CC}_p$, and $S^{C\mathrm{gt}}_p$ be the partial derivatives of $\mathrm{SSIM}_p$ with respect to these three statistics,
which have closed forms in the statistics at $p$, as in \texttt{fused-ssim}.
Since the three statistics depend on $C_q$ with the weights $\mathcal{W}(p-q)$, $2C_q\mathcal{W}(p-q)$, and $C^{\mathrm{gt}}_q\mathcal{W}(p-q)$,
and $\mathcal{W}$ is symmetric, the sum in \cref{eq:supp_fd_chain} becomes a convolution of the derivative maps:
\begin{equation}
\begin{split}
    g_q={}&\frac{1-\lambda}{M}\,\mathrm{sign}(C_q-C^{\mathrm{gt}}_q)
      -\frac{\lambda}{M}\Bigl[(\mathcal{W}*S^{C})_q\\
      &+2C_q(\mathcal{W}*S^{CC})_q+C^{\mathrm{gt}}_q(\mathcal{W}*S^{C\mathrm{gt}})_q\Bigr].
  \end{split}
  \label{eq:supp_fd_grad}
\end{equation}

\noindent\textbf{Kernels.}
F-D launches three kernels in sequence (\cref{alg:fd}).
We split the computation where a kernel needs results of other blocks,
since a kernel cannot wait for the other blocks of the same launch:
Kernel~2 needs the derivatives within 5 pixels of each tile, and Kernel~3 needs the sums over all blocks.
\begin{itemize}[leftmargin=1.2em,itemsep=1pt,topsep=2pt,parsep=0pt]
  \item \textbf{Kernel 1 (statistics).} Each block of threads handles one $16\times16$ tile of one channel.
  Since the $11\times11$ window of a pixel extends 5 pixels on each side,
  the block reads $C$ and $C^{\mathrm{gt}}$ of the tile and of a 5-pixel margin around it into shared memory once,
  computes the local statistics above,
  and writes the three derivative maps.
  The SSIM and L1 values are summed within the block and added to two global sums,
  instead of being written as maps.
  \item \textbf{Kernel 2 (gradient).} With the same tiling, it convolves the three derivative maps with $\mathcal{W}$
  and writes $g$ by \cref{eq:supp_fd_grad}, with $-\lambda/M$ and $(1-\lambda)/M$ passed as arguments instead of a map.
  \item \textbf{Kernel 3 (loss value).} A single thread forms $L$ from the two sums.
\end{itemize}
The only image-sized buffers are the three derivative maps and $g$, which need no zero initialization;
only the two sums are set to zero before Kernel~1.
Since the upstream gradient of $L$ is 1, the backward pass returns the stored $g$ without launching a kernel.
F-D thus replaces the 13 forward and 10 backward kernels of FastGS with three kernels in the forward pass and none in the backward pass.

\begin{algorithm}[t]
\caption{Fused loss with forward-pass gradient (F-D).}
\label{alg:fd}
\small
\begin{algorithmic}[1]
\Statex \textbf{Forward pass}
\State $\Sigma_{\mathrm{L1}}\gets0$, $\Sigma_{\mathrm{SSIM}}\gets0$ \Comment{two global sums}
\Statex \textbf{Kernel 1} (one block per $16\times16$ tile and channel)
\State load $C$, $C^{\mathrm{gt}}$ of the tile and a 5-pixel margin into shared memory
\State compute the five local statistics by separable $11\times11$ convolution
\State compute $\mathrm{SSIM}_p$ and $S^{C}_p$, $S^{CC}_p$, $S^{C\mathrm{gt}}_p$; store the derivatives
\State reduce $\lvert C_p-C^{\mathrm{gt}}_p\rvert$ and $\mathrm{SSIM}_p$ in the block; add them to $\Sigma_{\mathrm{L1}}$, $\Sigma_{\mathrm{SSIM}}$ atomically
\Statex \textbf{Kernel 2} (same tiling)
\State load the derivatives of the tile and the margin; convolve them with $\mathcal{W}$
\State compute $g_q$ by \cref{eq:supp_fd_grad} and store it
\Statex \textbf{Kernel 3} (one thread)
\State $L\gets(1-\lambda)\,\Sigma_{\mathrm{L1}}/M+\lambda\,(1-\Sigma_{\mathrm{SSIM}}/M)$
\Statex \textbf{Backward pass}
\State \Return the stored $g$ \Comment{no kernel}
\end{algorithmic}
\end{algorithm}

\noindent\textbf{Scope.}
F-D assumes that the loss is a plain mean of per-value terms and is back-propagated without weighting:
per-pixel weights would make $\partial L/\partial\ell_p$ no longer a constant,
and scaling the loss would require multiplying $g$ by the upstream gradient.
F-D also keeps one image-sized gradient from the forward pass to the backward pass.

\subsection{Implementation Optimizations for Compact Models (F-E)}
\label{sec:supp_fe}

F-E (Sec.~4.5 of the main paper) consists of four implementation optimizations built on those of Faster-GS~\cite{FasterGS}:
F-E3 and F-E4 adapt them to FastGS, and F-E1 and F-E2 also extend them.

%% ------------------------------------------------------------
\subsubsection{Fused Activations (F-E1)}
\label{sec:supp_fe1}

FastGS~\cite{FastGS} applies the sigmoid, the exponential, and the normalization to the raw opacity, scale, and rotation with PyTorch, which launches separate kernels and keeps intermediate tensors for the backward pass.
Following Faster-GS~\cite{FasterGS}, we pass the raw parameters to the rasterizer, whose kernels apply the activations and their derivatives:
$\sigma(1-\sigma)$ for the sigmoid $\sigma$, the exponential itself for the scale,
and $(\mathbf{I}-\hat{\mathbf{q}}\hat{\mathbf{q}}^\top)/\lVert\mathbf{q}\rVert$ for the normalization
$\hat{\mathbf{q}}=\mathbf{q}/\lVert\mathbf{q}\rVert$ of the rotation quaternion $\mathbf{q}$.

\noindent\textbf{Direct gradients (our extension).}
Faster-GS still returns the parameter gradients to automatic differentiation.
Our backward kernel instead adds them directly to the \texttt{.grad} buffers,
which removes the gradient tensors and the accumulation step of automatic differentiation.
It adds rather than overwrites, because FastGS steps the optimizer of some parameters only every several iterations,
and their gradients accumulate between the steps.

\begin{table}[t]
  \centering
  \footnotesize
  \setlength{\tabcolsep}{2pt}
  \caption{Parts of F-E1 in SprintGS-Base (averages over all 13 scenes). Time is in seconds.}
  \label{tab:supp_fe1}
  \begin{tabular}{lcccc}
    \toprule
    Activations and gradients & PSNR$\uparrow$ & SSIM$\uparrow$ & LPIPS$\downarrow$ & Time$\downarrow$ \\
    \midrule
    Pytorch (w/o F-E1, FastGS~\cite{FastGS})             & 27.38 & 0.822 & 0.252 & 91.1 \\
    In the rasterizer (Faster-GS~\cite{FasterGS})            & 27.38 & 0.822 & 0.252 & 84.3 \\
    \;+ direct gradients (ours, full F-E1)   & 27.38 & 0.822 & 0.252 & 80.0 \\
    \bottomrule
  \end{tabular}
\end{table}

\noindent\textbf{Effect.}
\Cref{tab:supp_fe1} separates the two parts of F-E1, keeping all other components and hyperparameters unchanged.
Applying the activations in the rasterizer, as in Faster-GS~\cite{FasterGS}, shortens the training from 91.1 to 84.3\,s,
and our direct gradients shorten it further to 80.0\,s,
accounting for 39\% of the gain of F-E1 at the same quality.

%% ------------------------------------------------------------
\subsubsection{Single-Launch Adam (F-E2)}
\label{sec:supp_fe2}

FastGS~\cite{FastGS} updates six parameter tensors with \texttt{torch.optim.Adam},
which launches several element-wise kernels per tensor.
Faster-GS~\cite{FasterGS} computes the Adam update of each tensor in one fused kernel.

\noindent\textbf{Single launch (our extension).}
We update all tensors of an optimizer in a single launch,
with a subclass of \texttt{torch.optim.Adam} whose \texttt{step} passes all tensors to one kernel
with the learning rate and the bias correction of each tensor.
The optimizer state keeps the layout of \texttt{torch.optim.Adam},
so the densification and pruning code of FastGS, which edits the state, works unchanged.

\begin{table}[t]
  \centering
  \footnotesize
  \setlength{\tabcolsep}{1pt}
  \caption{Parts of F-E2 in SprintGS-Base (averages over all 13 scenes). Time is in seconds.}
  \label{tab:supp_fe2}
  \begin{tabular}{lcccc}
    \toprule
    Adam update & PSNR$\uparrow$ & SSIM$\uparrow$ & LPIPS$\downarrow$ & Time$\downarrow$ \\
    \midrule
    Element-wise kernels (FastGS~\cite{FastGS})      & 27.37 & 0.822 & 0.252 & 84.3 \\
    Fused kernel per tensor (Faster-GS~\cite{FasterGS})       & 27.37 & 0.822 & 0.252 & 80.7 \\
    \;+ single launch (ours, full F-E2)       & 27.38 & 0.822 & 0.252 & 80.0 \\
    \bottomrule
  \end{tabular}
\end{table}

\noindent\textbf{Effect.}
\Cref{tab:supp_fe2} separates the two parts of F-E2 in the same way.
The fused kernel per tensor shortens the training from 84.3 to 80.7\,s, and the single launch saves a further 0.7\,s
by replacing the six launches per step with one.

\subsubsection{Densification Statistics in the Backward Kernel (F-E3)}
\label{sec:supp_fe3}

Densification decides which Gaussians to clone or split from per-Gaussian statistics
accumulated over the iterations, such as the norm of the gradient of the screen-space position.
FastGS~\cite{FastGS} updates these statistics after the backward pass with separate PyTorch operations,
which launch additional kernels and make the CPU wait for the GPU in every iteration.
Faster-GS~\cite{FasterGS} instead accumulates its densification statistics inside the backward kernel of the rasterizer,
where these gradients are computed.

\noindent\textbf{Our adaptation.}
SprintGS inherits the densification of FastGS, which uses four statistics:
the accumulated norms of the screen-space position gradient and of its absolute counterpart,
the number of views in which the Gaussian is visible, and its maximum screen-space radius.
Since these statistics differ from those of Faster-GS,
we adopt the idea of Faster-GS and apply it to the four statistics of FastGS:
the backward kernel updates them directly, with the same definitions as in FastGS.
The densification of FastGS, including its thresholds, is thus used unchanged,
and the separate update after the backward pass is no longer needed.

%% ------------------------------------------------------------
\subsubsection{Asynchronous Training Loop (F-E4)}
\label{sec:supp_fe4}
The training loop of FastGS~\cite{FastGS} synchronizes the CPU with the GPU several times per iteration to read the loss for the progress display and query CUDA-event timings.
These waits delay subsequent kernel launches and can introduce GPU idle time. Following Faster-GS~\cite{FasterGS}, we defer these reads to
the progress-update intervals.

\noindent\textbf{Our adaptation.}
We retain the progress display and its CUDA-event timing
statistics while removing the per-iteration waits. The running loss is kept on the GPU, and four CUDA events
per iteration are recorded into a pool. Every 500 iterations and at the end of training, the CPU synchronizes once, sums the elapsed times of the pooled event pairs, and reads the loss. These CUDA-event timings are used only for the progress display. All training times reported in the paper are measured separately with the same host-side wall-clock timer for all methods, with CUDA synchronization at both ends of the measured training loop, as described in Sec.~5.1 of the main paper.
Data loading, evaluation, logging, and saving are excluded.

\end{document}